\documentclass[10pt,twocolumn,letterpaper]{article}

\usepackage[pagenumbers]{iccv} 

\usepackage{multirow}
\usepackage{xspace}
\usepackage{amsmath, amssymb, algorithm, algpseudocode, graphicx}

\newcommand{\eazy}{\texttt{EAZY}\xspace}
\DeclareMathOperator{\olvar}{OL-VAR}
\DeclareMathOperator{\chair}{CHAIR}
\DeclareMathOperator{\attn}{Attention}
\DeclareMathOperator{\smax}{softmax}

\usepackage{booktabs}
\usepackage{times}
\usepackage{epsfig}
\usepackage{pifont}
\usepackage{colortbl}
\usepackage{multirow}
\usepackage{diagbox}

\usepackage{svg}
\usepackage{wrapfig,lipsum,booktabs}

\definecolor{iceblue}{RGB}{33,102,200}
\definecolor{fired}{RGB}{222,82,57}
\usepackage[export]{adjustbox}
\definecolor{mygray}{gray}{.9}
\definecolor{mygreen2}{RGB}{80,100,40}  
\definecolor{mygreen3}{RGB}{80,99,43} 
\definecolor{myred}{RGB}{147,58,50} 
\definecolor{myyellow}{RGB}{255,192,0} 
\definecolor{OliveGreen}{RGB}{50,160,0}

\usepackage{caption}

\definecolor{iccvblue}{rgb}{0.21,0.49,0.74}
\usepackage[pagebackref,breaklinks,colorlinks,allcolors=iccvblue]{hyperref}

\def\paperID{2933} 
\def\confName{ICCV}
\def\confYear{2025}

\title{Hallucinatory Image Tokens: A Training-free \eazy Approach on Detecting and Mitigating Object Hallucinations in LVLMs}

\author{Liwei Che\\
{\tt\small lc1279@cs.rutgers.edu}
\and
Tony Qingze Liu\\
{\tt\small ql236@scarletmail.rutgers.edu}
\and
Jing Jia\\
{\tt\small jing.jia@rutgers.edu}
\and
Weiyi Qin\\
{\tt\small wq50@cs.rutgers.edu}
\and
Ruixiang Tang\\
{\tt\small ruixiang.tang@rutgers.edu}
\and
Vladimir Pavlovic\\
{\tt\small vladimir@cs.rutgers.edu}
\and
Rutgers University\\
New Brunswick, NJ, USA 08901\\
}

\begin{document}
\maketitle

\begin{abstract}
    Despite their remarkable potential, Large Vision-Language Models (LVLMs) still face challenges with object hallucination, a problem where their generated outputs mistakenly incorporate objects that do not actually exist.
   Although most works focus on addressing this issue within the language-model backbone, our work shifts the focus to the image input source, investigating how specific image tokens contribute to hallucinations. Our analysis reveals a striking finding: a small subset of image tokens with high attention scores are the primary drivers of object hallucination. By removing these hallucinatory image tokens (only $1.5\%$ of all image tokens), the issue can be effectively mitigated. This finding holds consistently across different models and datasets. Building on this insight, we introduce \eazy, a novel, training-free method that automatically identifies and \textbf{E}liminate h\textbf{A}llucinations by \textbf{Z}eroing out hallucinator\textbf{Y} image tokens. We utilize \eazy for unsupervised object hallucination detection, achieving $15\%$ improvement compared to previous methods. Additionally, \eazy demonstrates remarkable effectiveness in mitigating hallucinations while preserving model utility and seamlessly adapting to various LVLM architectures.
\end{abstract}

\begin{figure*}[!h]
    \centering
    \includegraphics[width=\linewidth]{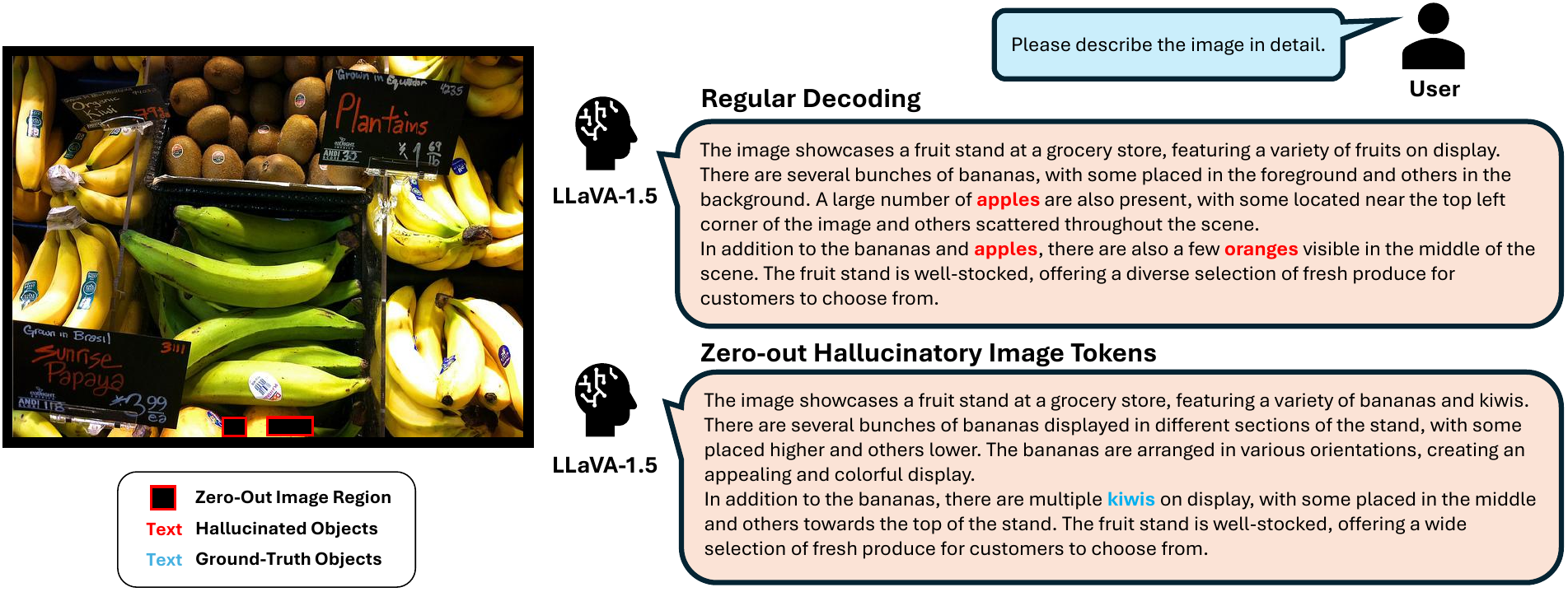}
    \caption{Removing three image tokens results in the elimination of the hallucinated objects, "apples" and "oranges", and reveals the real object "kiwis".}
    \label{fig:case_study}
\end{figure*}

\section{Introduction}

Large Vision-Language Models (LVLMs)~\cite{liu2024visual,liu2024improved,chiang2023vicuna} have achieved remarkable advancements, seamlessly integrating visual recognition and language understanding to produce outputs that are both coherent and contextually aligned. Despite these successes, LVLMs face a critical challenge from hallucination~\cite{liu2024survey}. Hallucination in LVLMs manifests as generating incorrect responses that either contradict the provided image input or deviate from user instructions. This problem not only compromises LVLMs reliability but also limits their broader applicability in real-world scenarios. Among the various forms of hallucination, object hallucination (OH)~\cite{rohrbach2018object} is particularly prevalent. This occurs when LVLMs either misidentify objects or incorrectly perceive hallucinatory objects (HO) as being present in an image. Addressing this issue is essential to enhance the accuracy and trustworthiness of LVLMs for practical applications.


Various studies~\cite{rohrbach2018object,sun2023aligning,guan2024hallusionbench} have been conducted to investigate the underlying causes of OHs in LVLMs. A significant portion of existing research attributes this issue to the biases from the language part. In~\cite{li2023evaluating,leng2024mitigating,zhou2023analyzing}, the authors argue that statistical biases in the textual training data are a major contributing factor. These biases include the long-tailed distribution of object occurrence frequencies~\cite{li2023evaluating} and the co-occurrence relationships between different objects~\cite{rohrbach2018object,zhou2023analyzing}.
Some other works~\cite{leng2024mitigating,lee2023volcano,wang2023evaluation} believe that prior knowledge from the powerful language model overrides input information, resulting in various hallucinations. Consequently, these studies employ contrastive decoding~\cite{li2022contrastive}, which achieves decoding correction by comparing token probability distributions before and after modification to image input.
Another line of research noticed that the language model pays less attention to the image tokens but more to the most recent text token during the generation process therefore applying generation backtracking~\cite{huang2024opera,wei2024dopra}.  
While these pioneering efforts have provided valuable insights into the causes of visual hallucinations, a key question remains: \textit{How does visual input influence the generation of visual hallucinations?} This study aims to explore this critical aspect in depth to better understand hallucination phenomena from the vision perspective.


To address this gap, we investigated how LVLMs process visual inputs and how this contributes to visual OHs. 
Although using the distribution of attention weights to interpret visual token contributions is a straightforward approach~\cite{neo2024towards,gong2024damro,woo2024don}, many layers exhibit noisy attention patterns with no clear correlation between generated object-related tokens and specific image regions.
Through a comprehensive analysis of different layers, as shown in Figure~\ref{fig:layer_attention}, we found that LLaVA and similar models predominantly extract object information in their middle-to-late layers. 
For example, in LLaVA-1.5 (7b), a 32-layer transformer model, we reveal that in the middle layers, such as Layer 15, the generated object tokens assign a large proportion of their attention weights to their visual grounding regions, as shown in Figure~\ref{fig:bbox_ratio}. 
This phenomenon effectively links generated object-related tokens to precise regions in the input image and is notably absent in the earlier and final layers. This insight allows us to pinpoint where the model “looks” in the input image while generating text tokens. 



Equipped with this analytical tool, we found that real objects tend to focus their attention on their corresponding image anchor regions. In contrast, hallucinatory objects, lacking a true image anchor, concentrate their attention on visually related regions. For example, a nonexistent \textit{baseball glove} could focus its attention on the \textit{baseball} in the image. 
This finding highlights the critical role of visual bias in generating object hallucinations. When the image anchor of a real object is masked, the model fails to recognize the object. Inspired by this, we apply a \textit{zero-out} operation to the top-$K$ image tokens with the highest attention scores for hallucinated objects, replacing them with zero embeddings. Surprisingly, these hallucinated objects disappear from the new response. 
We further identified certain image tokens among the top-$K$ zeroed-out tokens—termed \textbf{Hallucinatory Image Tokens (HITs)}—that play a key role in generating object hallucinations. As shown in Figure~\ref{fig:case_study}, the hallucinated object could be eliminated by removing even two or three HITs. 
In contrast, we found that real objects remain largely unaffected by the zero-out operation.
This distinction shows a promising way to both detect and mitigate object hallucination.

To evaluate the generalizability of this finding, we curated a dataset of hallucination cases and applied the top-$K$ candidate HITs zero-out operation. 
The results show that we can successfully reduce most OHs while preserving accurate predictions for real objects.
We extended these findings and proposed a novel method, \eazy, which can detect and \textbf{E}liminate object h\textbf{A}llucination by \textbf{Z}eroing-out hallucinator\textbf{Y} image tokens. \eazy first identifies hallucinated objects in the generated text by removing object-specific top-$K$ candidate HITs, with a detection rate up to $80\%$. After the hallucination detection stage, the \eazy will rectify the generated response by zeroing out the aggregated HIT candidates from all the detected HOs. Our experiment results show that \eazy surpassed the performance of state-of-the-art (SOTA) training-free methods.





Our contributions can be summarized as below:

\begin{itemize}
    \item We show that the LVLMs extract object-related information from the visual input in the early middle to late middle layers. This relationship builds strong connections between visual regions and generated object tokens, allowing object localization via attention distribution.

    \item We discovered an interesting and significant phenomenon: Most HOs generated by LVLMs are directly linked to only a subset of image tokens receiving high attention scores. By zeroing out these \textit{hallucinatory image tokens}, the associated HOs can be eliminated.

    \item By further validation of the HITs phenomenon, we found that zeroing out top-$K$ candidate HITs not only removes the majority of HOs but has minimal impact on real objects. We propose to use the HITs phenomenon for OH detection, achieving nearly $80\%$ accuracy and precision for both real and hallucinatory objects. 
    
    \item We propose a method named $\eazy$ for detecting and mitigating OH by zeroing out candidate HITs. Our experimental results demonstrate that the proposed method can adapt to various LVLMs and outperforms state-of-the-art baseline approaches across most evaluation metrics.
\end{itemize}



\section{Connecting Generated Object Token with Its Image Anchor via Attention}
\label{sec:localize}

\noindent We consider LVLMs represented by LLaVA\cite{liu2024visual}, with a model architecture consisting of a vision encoder, a vision-text connecting module, and a transformer-based large language model as the text decoder. An input image $I$ is first divided into $N$ fixed-size patches and then processed by ViT $f_V$ and a linear projector or MLP  into a sequence of image tokens. For LLaVA-1.5, a single image will be mapped as $576$ image tokens for the LLM decoder.

Given an input sequence $\mathbf{x} = (x_0, x_1, \ldots, x_{M+N-1})$, the text decoder $p(x_t|\mathbf{x}_{<t}) = f_T(\mathbf{x}_{<t}|\theta)$ gives the probability distribution of next token over the vocabulary space. Here, $M$ is the text token number; $\theta$ are the model parameters; $\mathbf{x}_{<t}$ is the input sequence that contains the tokens prior to the $t$-th token $x_t$. The complete generated token sequence $\mathbf{x}_{output}$ is obtained via $F_T(\mathbf{x}|\theta)$. We denote $\mathbf{x}_{V}$ as input image tokens and $\mathbf{x}_{T}$ as the input text tokens. 

For a token $x_i$, its representation $h^l_i \in \mathcal{R}^{d_m}$ at the layer $l$ of the text decoder is updated through multi-head self-attention (MHSA) and feed-forward(FF) sublayers with residual connection. Specifically, for the MHSA sublayer with $H$ heads, each head applies self-attention as:
\begin{equation}
    \attn(Q,K,V) = \smax(\frac{QK^T}{\sqrt{d_h}}+M)V
\end{equation}
Here $Q,K,V \in \mathcal{R}^{n\times d_k}$ are the query, key, and value matrices linearly projected from the input; $M \in \mathcal{R}^{n\times n}$ is the casual mask. We utilize the average softmax output of the self-attention $a^l = \smax(\frac{QK^T}{\sqrt{d_h}}+M)$ as the attention scores at layer $l$. We denote the casual self-attention scores from the token $x_i$ to $x_j$ at layer $l$ as $a^l_{ij}$.

\subsection{Evaluating Object Localization Performance among Model Layers.}


\label{sec:locate_image_tokens}

\begin{figure}[!ht]
    \centering
    \begin{minipage}{0.49\linewidth}
        \includegraphics[width=\textwidth]{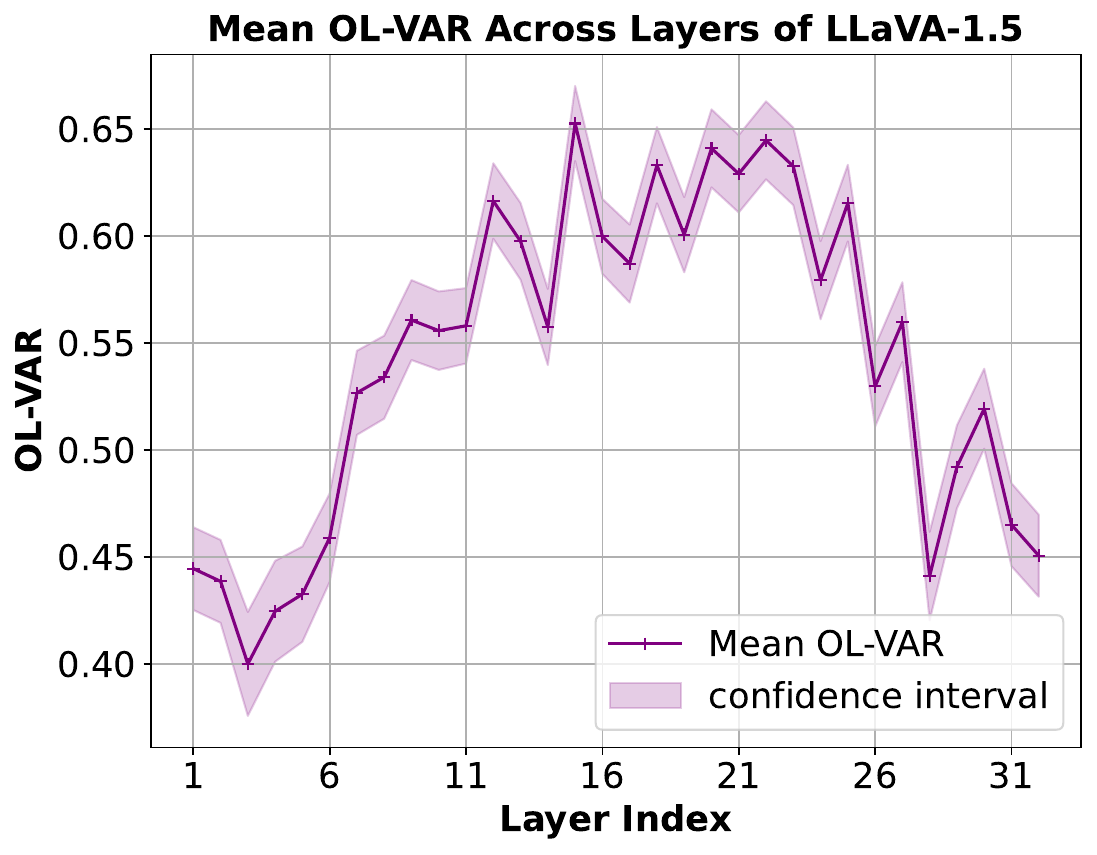} 
    \end{minipage}
    \begin{minipage}{0.49\linewidth}
        \centering
        \includegraphics[width=\textwidth]{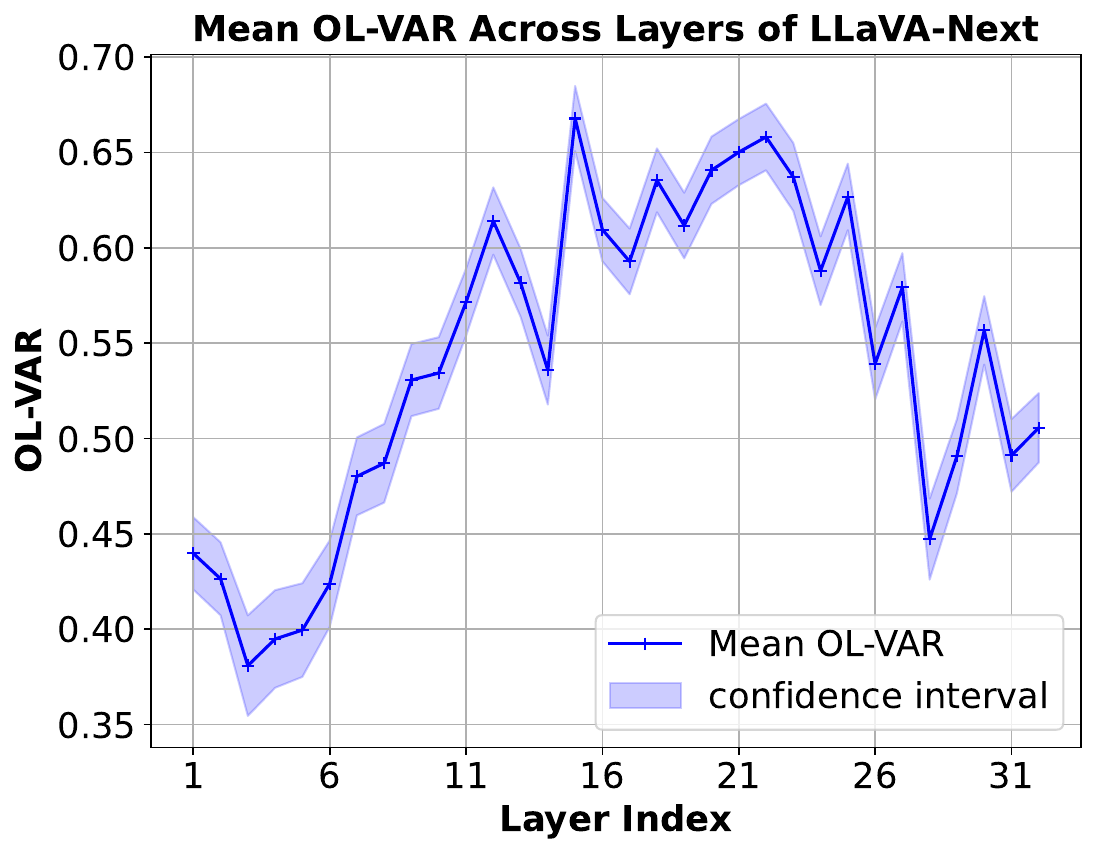}
    \end{minipage}
    \caption{Average $\olvar$ over $500$ randomly selected MSCOCO images of \text{LLaVA-1.5}(left) and LLaVA-Next(right).} 
    \label{fig:bbox_ratio} 
\end{figure}

\begin{figure}[!ht]
    \centering
    \includegraphics[width=\linewidth]{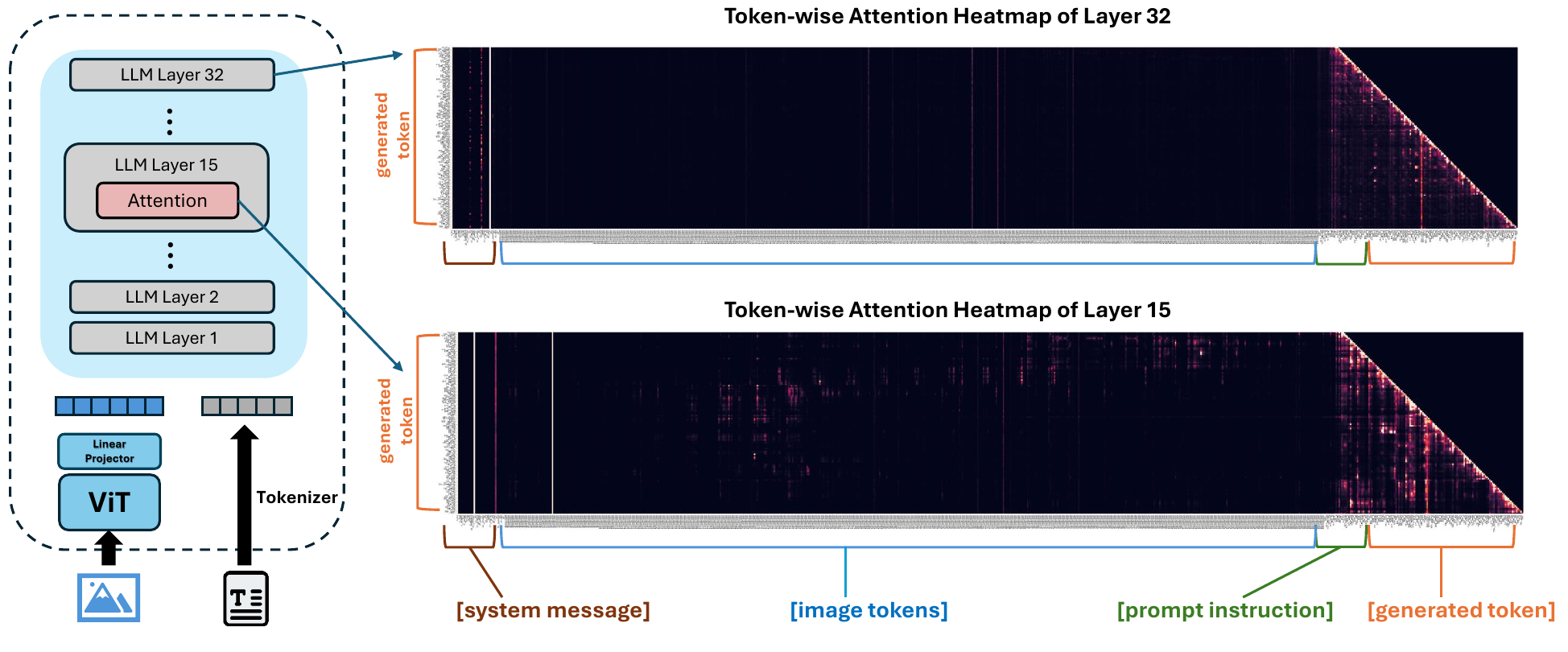}
    \caption{The token-wise attention heatmap of layer 32 and 15 from the LLaVA model. The attention distribution in layer 15 illustrates the connection between the generated object tokens and image tokens, indicating the model extracts object information from the image at the middle layers.}
    \label{fig:layer_attention}
\end{figure}

Understanding how LVLMs process visual input is still in its early stages. 
Recent studies~\cite{gong2024damro,woo2024don} have preliminarily revealed that models such as LLaVA assign a higher proportion of attention weights over the image tokens in their middle to late layers.
However, these analyses fail to identify a clear dependency between the generated object-related tokens and specific image regions within the cluttered, layer-by-layer attention distributions.


To explore where the model establishes connections between the generated object text tokens and their corresponding image regions, i.e., image anchors, we evaluated the performance of object localization using token-to-token attention relationships at each layer. 
To quantitatively assess the relationship between generated object tokens and their corresponding image regions, we introduced a new metric, the Object Localization Visual Attention Ratio (OL-VAR), to represent the ratio of attention scores over the object bounding box region to all the image tokens by generated object token $x_i$ in layer $l$:
\begin{equation}
    \olvar^l_{i} = \frac{\sum_{x_k\in \mathbf{x}_{V_O}} a^l_{ik}}{\sum_{x_j\in \mathbf{x}_{V}} a^l_{ij}}.
\end{equation}
Here, $\mathbf{x}_{V}$ represents the set of all the image tokens and $ \mathbf{x}_{V_O}$ is the set of image tokens contained in the ground truth bounding box of the target object $O$. We randomly selected $500$ MSCOCO images containing annotated objects and their corresponding bounding boxes and show their average $\olvar$ over each layer of LLaVA-1.5 and LLaVA-Next in Figure~\ref{fig:bbox_ratio}. It can be seen that the model assigns more than $55\%$ of the visual attention weights on the anchors of the object-related image from layer $10$ to $25$, where layer $15$ obtained the highest score $\olvar$. This result offers a new perspective on understanding how LVLMs extract object information from visual inputs, revealing that the model primarily extracts and comprehends object information in the early-middle to later-middle layers. In addition, it provides an effective analytical tool for establishing connections between visual inputs and generated tokens.




We further validated this finding through the visualization of attention heatmaps. In Figure~\ref{fig:layer_attention}, we present two representative token-wise attention heatmaps of LLaVA-1.5 7b with $32$ transformer layers. The top panel shows the attention distribution for Layer $32$, while the bottom panel depicts the distribution for Layer $15$, the layer with the highest $\olvar$ on objects. 
For any given image input, the attention heatmaps of most layers exhibit patterns similar to those of Layer 32. Specifically, the attention scores from generated tokens to the image tend to concentrate on a few specific image tokens, forming an \textit{attention sink} pattern \cite{xiao2023efficient}.
In contrast, the attention heatmap of Layer $15$ exhibits a markedly different distribution. The amount of attention sink pattern is significantly diminished. Instead, multiple distinct attention spots emerge, connecting the generated text tokens with various image tokens. This indicates a more distributed attention mechanism at layer $15$, enabling richer interactions between text and visual elements.

\section{Object Hallucination from Image Tokens.} 
\label{sec:visual_bias}


This section presents a new perspective on the causes of hallucinations. We found that the presence of hallucinated objects in the generated text is directly correlated with the few image tokens that receive the highest attention scores. We illustrate more details about the distribution of text-to-image token attention in Appendix~\ref{sec:real_HO_heatmap}.



\subsection{Identify Hallucinatory Image Tokens}
\label{sec:hits}


\noindent We begin with a case study where the model is prompted to describe an image of \textit{``a fruit stand with bananas and kiwis"}, as shown in Figure~\ref{fig:case_study}. The discovery in Section~\ref{sec:localize} provides a tool for connecting generated object tokens with their visual grounding. Erasing the image anchor region of a real object prevents the model from recognizing its presence. Inspired by this, we selected the top five image tokens that received the highest attention scores for the hallucinatory objects "apple" and "orange" in the original generation and replaced them with zero embeddings ($14$ image tokens from multiple HO tokens). Surprisingly, "apple" and "orange" disappeared, while the real object "kiwi" emerged.

\begin{figure}[!ht]
    \centering
    \includegraphics[width=\linewidth]{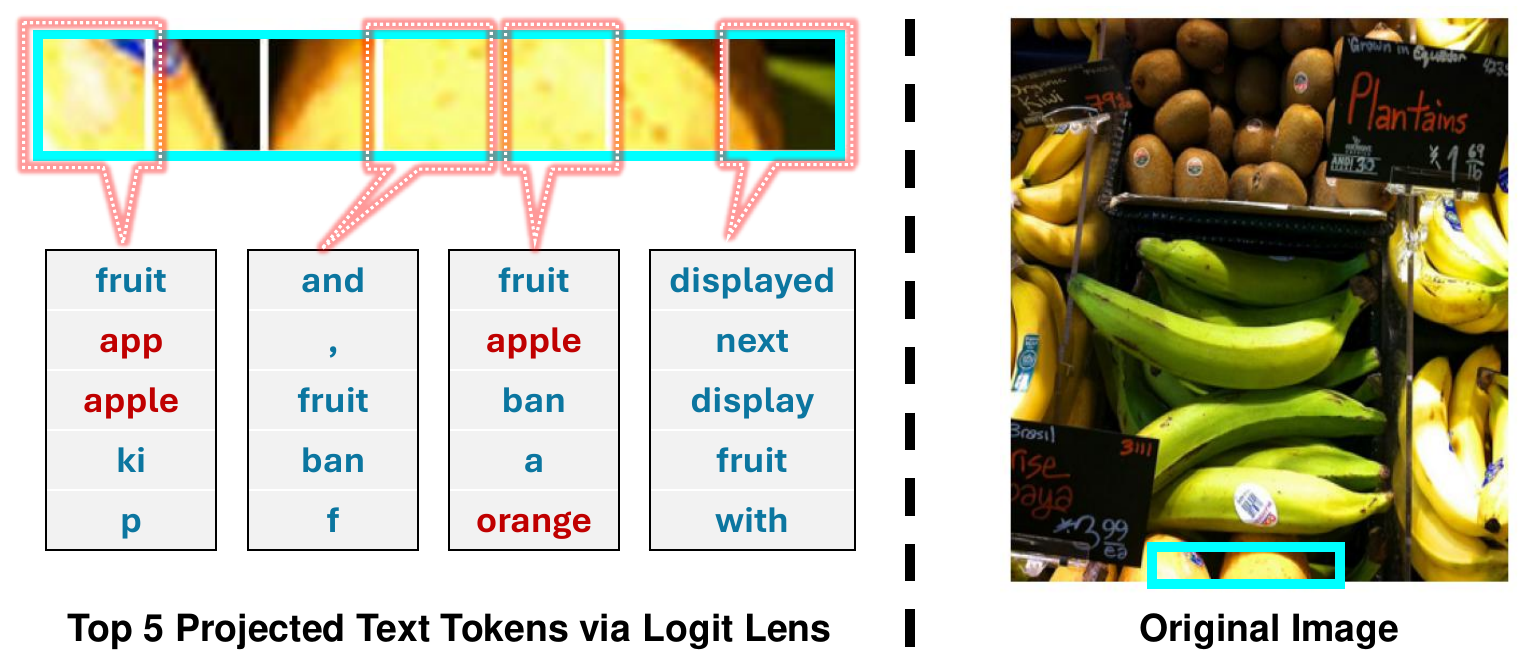}
    \caption{Logit Lens~\cite{nostalgebraist2020logitlens} interpretation of HITs. We display the top-$5$ projected words of image tokens related to HOs. The HITs are interpreted as the hallucinatory objects, "apple" and "orange".}
    \label{fig:logit_lens}
\end{figure}


Our further analysis aims to identify the smallest set of image tokens capable of eliminating hallucinations. Ultimately, we identified three key image tokens among the top-attended tokens, located in the lower part of the 'papaya' region. Remarkably, zeroing out just these three tokens achieved the same effect as removing all 14 image tokens with high attention scores.
To further investigate how the LLM interprets these image tokens, we applied the Logit Lens~\cite{nostalgebraist2020logitlens} technique, which maps the hidden states of image tokens to the LLM's vocabulary space. In Figure~\ref{fig:logit_lens}, we show the top-$5$ projected vocabulary words associated with several image tokens in the papaya region. Surprisingly, the three key image tokens are interpreted as "apple" and "orange" by Logit Lens. 
This aligned with our initial observation that certain image tokens are directly associated with hallucinatory objects because of inherent visual biases. We define this type of image tokens as the \textit{\textbf{Hallucinatory Image Tokens}} (HITs)\footnote{Logit Lens alone cannot capture all HITs. 
Interpretability-based methods for detecting such tokens could offer improved alignment, and we plan to consider alternate tools to refine this approach in the future.}.
In the next section, we demonstrate that zeroing out the top-$K$ candidate HITs (top-$K$ most attended image tokens by HOs) can universally reduce object hallucination across different images and contexts.

\begin{figure}[h!]
    \centering
    \begin{minipage}{0.49\linewidth}
        \includegraphics[width=\textwidth]{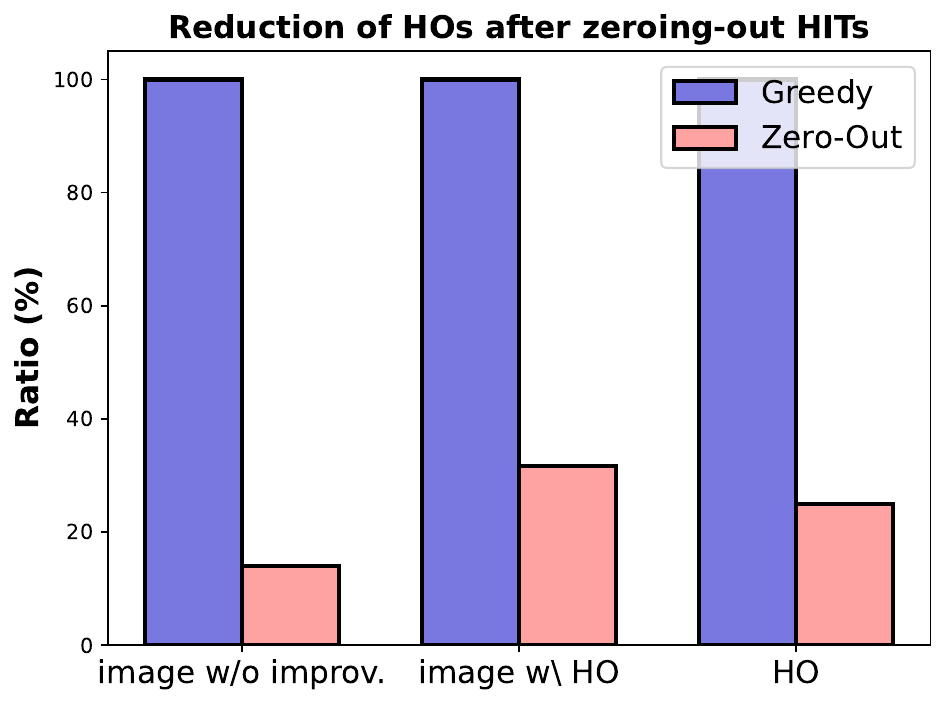}
    \end{minipage}
    \begin{minipage}{0.49\linewidth}
        \includegraphics[width=\textwidth]{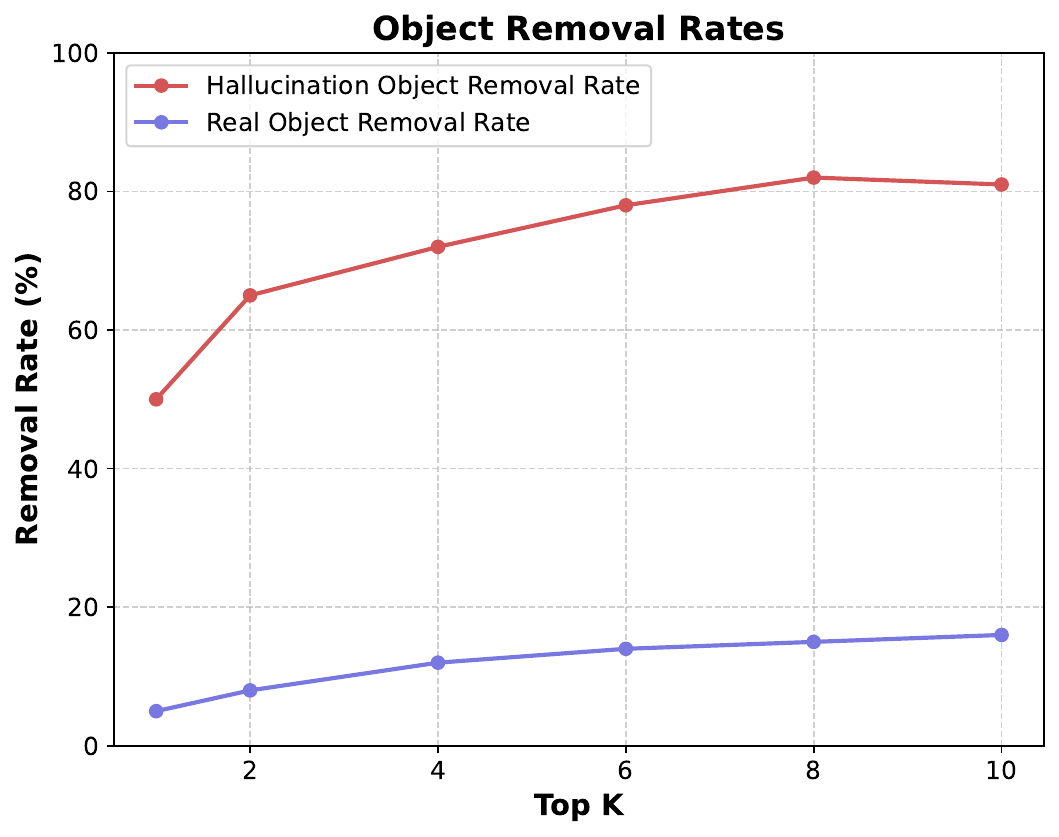}
    \end{minipage}
    \caption{\textbf{Left:} Reduction ratio of OHs in Hall-COCO after zeroing-out HITs. We show the ratio changes of image w/o improv.(ratio of images with the same HOs after zeroing out), image w\ OH (images with OH in the response), and HO (Ratio of HOs in all text responses). \textbf{Right:} The fraction change of real and hallucinatory objects removed by zeroing out the top-$K$ candidate HITs.} 
    \label{fig:removal} 
\end{figure}

\subsection{Quantifying the Impact of Zeroing Out HITs}
\label{sec:universal_effect}

\noindent To further validate that the HITs phenomenon is broadly applicable, we curated a dataset named \textit{Hall-COCO}, containing $200$ images from MSCOCO dataset~\cite{rohrbach2018object} that consistently induce object hallucinations identified by human annotation. We provide more details in Appendix~\ref{sec:hallcoco}.

We evaluate how zero-out the candidate HITs (top attended image tokens by HO tokens) will impact the number of object hallucinations in the model's responses.
As presented in Figure~\ref{fig:removal} Left, we can see that after zero-outing the top 5 candidate HITs with the same greedy decoding, $86\%$ of images got improved description with less hallucination. The proportion of images with hallucination decreased from $100\%$ to $31.73\%$. $74.17\%$ of the object hallucination directly disappeared from the updated generation. This verified that most hallucinatory image tokens are part of the image tokens that received the most attention weights from the OH tokens. Removing these top-$K$ candidate hallucinatory image tokens leading to the removal of OH can be widely applied to different images.

\subsection{Real Objects are Robust to Zero-Out }
\label{sec:obj_robust}

\noindent We demonstrate that zeroing out hallucinatory image tokens significantly reduces the generation of hallucinated object tokens. This raises an interesting question: What is the effect of zeroing out high-attention patches for real objects? To explore this, we evaluate the impact of zeroing out the top-$K$ HITs on both hallucinated and real objects. Specifically, we measure the fraction of objects removed by counting how many objects remain in the model response after removing the top-$K$ image tokens with the highest attention scores from the corresponding text object token, relative to the total number of objects. The results, shown in Figure~\ref{fig:removal} (Right), reveal that object hallucinations are highly sensitive to zeroed HITs. For example, when $K = 8$, approximately 80\% of hallucinated objects were successfully eliminated, while only a small fraction (~10\%) of real objects were affected. Notably, the few real objects that were removed tended to be small or positioned at the periphery of the image. We provide a failure case study in Appendix~\ref{sec:failure_Case}.

\section{Detecting and Mitigating Object Hallucination by HIT Zero-outs}

In this section, we introduce our proposed methods, \eazy (\underline{E}liminate object h\underline{A}llucination by \underline{Z}ero-out hallucinator\underline{Y} image tokens), an automatic approach to detect and mitigate object hallucinations in LVLMs. 

\begin{figure*}
    \centering
    \includegraphics[width=0.8\linewidth]{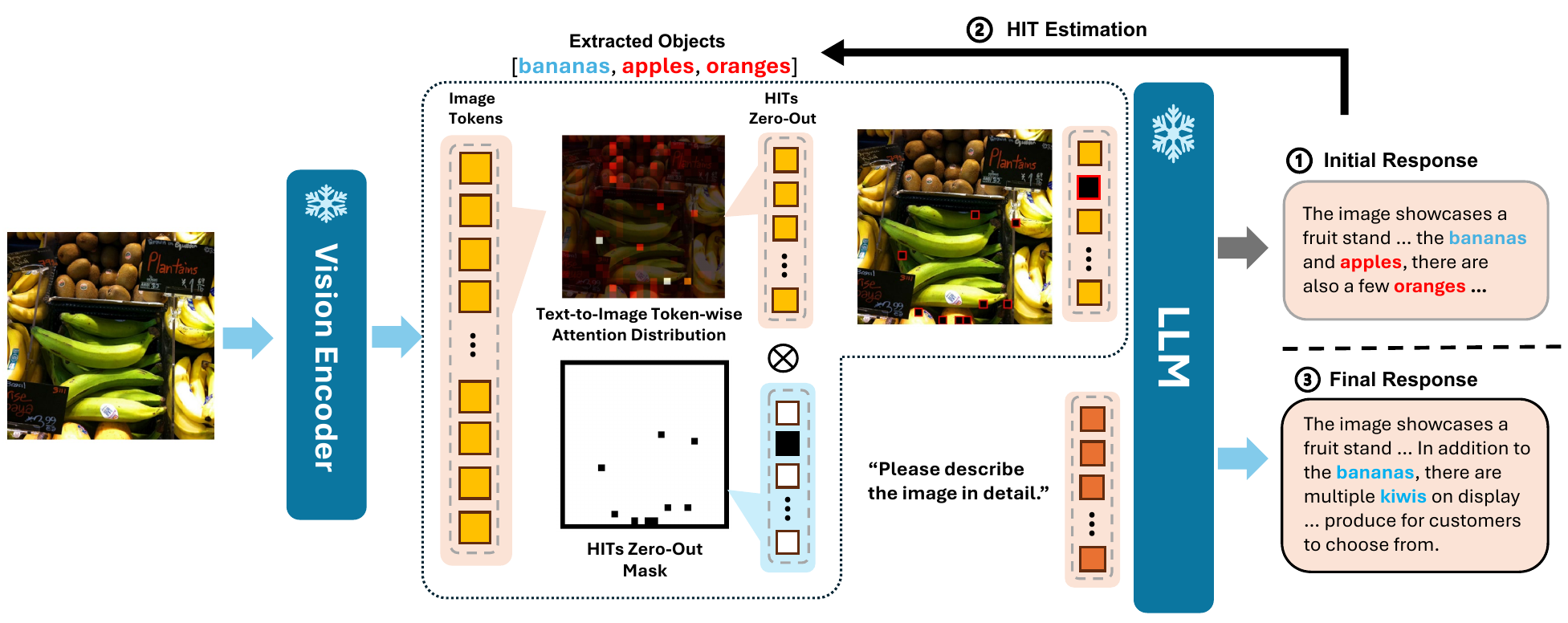}
    \caption{Overview of the proposed \eazy method. The process starts with an image input encoded into image tokens. The LVLM generates an initial response with hallucinated objects (\textcolor{red}{apples}, \textcolor{red}{oranges}). \eazy estimates HITs via text-to-image token-wise attention distribution. With HITs zeroed out, the final response has the hallucinations disappeared, revealing correct objects (\textcolor{blue}{kiwis}).}
    \label{fig:eazy}
\end{figure*}

\subsection{Object Hallucination Detection}
Our results in Section~\ref{sec:localize} and~\ref{sec:visual_bias} indicate that removing the top-$K$ candidate HITs from consideration effectively eliminates HOs in the revised model output, while largely preserving the integrity of real objects. Leveraging this differentiation between genuine and HO tokens, we suggest identifying OH by discarding the top-$K$ candidate HITs.

Given an image input $I$ and its corresponding input token sequence $\mathbf{x}_{I}$, the text decoder produces $G$ generated tokens $\mathbf{x}_{output} = \{x_{M+N}, \ldots, x_{M+N+G-1}\}$. We then utilize pre-built NLP techniques such as POS tagging and dependency parsing~\cite{bird2006nltk} to extract object tokens in the generated response $\mathbf{x}_{output}$, resulting in a subset $\mathbf{x}_{obj}$. For any $x_i \in \mathbf{x}_{obj}$, let $a_{ij}$ represent the attention score of $x_i$ on an image token $x_j$. The zero-out token set $\mathbf{z}_i$ is defined as
\begin{equation}
    \mathbf{z}_i = \left\{ x_j \in \mathbf{x}_{I} \;\middle|\; j \in \operatorname{arg\,top}_K\left(\{a_{ij} \mid x_j \in \mathbf{x}_{I}\}\right) \right\},
\end{equation}
where $\operatorname{arg\,top}_K$ selects the indices of the $ K $ largest attention scores $\{a_{ij} \mid x_j \in \mathbf{x}_{I}\}$. We modify the image input by applying zero-out based on $\mathbf{z}_i$,
\begin{equation}
    \mathbf{x}^{\mathbf{z}_i}_{input} = \left\{ x_j \leftarrow \mathbf{0}_{d_m} \;|\; x_j \in \mathbf{z}_i \right\}.
\end{equation}
Here, $ \mathbf{x}^{\mathbf{z}_i}_{\text{input}} $ indicates the input sequence with the tokens in $ \mathbf{z}_i $ replaced by zeroing embedding $\mathbf{0}_{d_m} \in \mathcal{R}^{d_m}$.
We check the updated generation result $\mathbf{x}_{output}^i$ by text decoder $\mathbf{x}_{output}^i=F_T(\mathbf{x}^{\mathbf{z}_i}_{\text{input}} \mid \theta)$. If $ x_i \notin \mathbf{x}^i_{\text{output}} $, \eazy classifies $ x_i $ as an object hallucination, and a real object otherwise.


\subsection{Mitigating Object Hallucination}

\begin{algorithm}[t]
    \caption{\eazy}
    \label{alg:eazy}
    \begin{algorithmic}[1]
        \Require input tokens $\mathbf{x}_{input}$, text decoder $F_T$, attention scores $a_{ij}$, top-$K$ threshold $K$, input image $I$
        \Ensure Mitigated response $\mathbf{x}_{output}$
        
        \State Compute initial response: $\mathbf{x}_{output} \gets F_T(\mathbf{x}_{input} \mid \theta)$
        \State Extract object tokens: $\mathbf{x}_{obj} \gets \mathbf{x}_{output}$
        
        \State Identify top-$K$ attended image tokens for all objects:
        \begin{equation*}
            \mathcal{Z}^{estimate}_{I} = \bigcup_{x_i \in \mathbf{x}_{obj}} \operatorname{arg\,top}_K(\{a_{ij} \mid x_j \in \mathbf{x}_{obj}\})
        \end{equation*}
        
        \State Zero-out all hallucinated tokens: $\mathbf{x}^{\mathcal{Z}^{estimate}_I}_{input} \gets \mathbf{x}_I$ with $\mathcal{Z}^{estimate}_I$ replaced by $\mathbf{0}_{d_m}$
        \State Generate updated response: $\mathbf{x}_{output}^{HIT} = F_T(\mathbf{x}^{\mathcal{Z}_I}_{input} \mid \theta)$
        
        \State Identify hallucinated objects: $\mathbf{x}^{HO}_{obj} = \{x_i \in \mathbf{x}_{obj} \mid x_i \notin \mathbf{x}_{output}^{HIT}\}$
        
        \State Final zero-out of hallucinated tokens:
        \begin{equation*}
            \mathcal{Z}^{final}_I = \bigcup_{x_i \in \mathbf{x}_{halluc}} \operatorname{arg\,top}_K(\{a_{ij} \mid x_j \in \mathbf{x}_I\})
        \end{equation*}
        
        \State Generate final response: $\mathbf{x}^{\eazy}_{output} = F_T(\mathbf{x}^{\mathcal{Z}^{final}_I}_{input} \mid \theta)$
        
        \Return $\mathbf{x}^{\eazy}_{output}$
    \end{algorithmic}
\end{algorithm}


As shown in Figure~\ref{fig:eazy}, the mitigation application of \eazy starts from an initial inference without intervention. Following a similar process as the detection application, the object tokens $x_{obj}$ are extracted from the initial response. We then group the top-$K$ candidate HITs $\mathcal{Z}^{estimate}_I$ for all objects $x_i \in \mathbf{x}_{obj}$ and apply zero-out to the image tokens accordingly. The model then performs inference for HITs estimation after zeroing out the corresponding image tokens. \eazy identifies objects that disappear in the new response as hallucinatory object tokens $\mathbf{x}^{HO}_{obj}$.

We collect all the top-$K$ candidate HIT sets of the HO tokens detected from the previous HIT estimate stage. We take the union set of all the candidate HITs sets to obtain the final zero-out token list $\mathcal{Z}^{final}_I$, which is defined as
\begin{equation}
    \mathcal{Z}^{final}_I = \bigcup_{x_i \in \mathbf{x}_{obj}} \mathbf{z}_i \quad \text{where} \quad x_i \in \mathbf{x}^{HO}_{obj}.
\end{equation}
The model then utilizes the final zero-out token list $\mathcal{Z}^{final}_I$ to replace all the candidate hallucinatory image tokens by zero embedding for the final generation:
\begin{equation}
    \mathbf{x}^{\eazy}_{output} = F_T(\mathbf{x}^{\mathcal{Z}^{final}_I}_{input} \mid \theta).
\end{equation}

\section{Experiment}

In this section, we conduct experiments on multiple datasets and metrics to validate the superior performance of \eazy in detecting and mitigating hallucinated objects (HOs).

\begin{table*}[!htbp]
    \centering
    \small
    \begin{tabular}{l cc cc cc}
        \toprule
        \multirow{2}{*}{\textbf{Method}} & \multicolumn{2}{c}{\textbf{LLaVA-1.5}} & \multicolumn{2}{c}{\textbf{Shikra}} & \multicolumn{2}{c}{\textbf{LLaVA-NEXT}} \\
        \cmidrule(r){2-3} \cmidrule(l){4-5} \cmidrule(l){6-7}
        & $\text{CHAIR}_S \downarrow$ & $\text{CHAIR}_I \downarrow$ &  $\text{CHAIR}_S \downarrow$ & $\text{CHAIR}_I \downarrow$
        &  $\text{CHAIR}_S \downarrow$ & $\text{CHAIR}_I \downarrow$\\
        \midrule
        Greedy & $49.6$ & $14.4$ & $55.8$& $15.4$ & $32.8$& $9.1$ \\
        Beam Search & $46.3$ & $12.9$ & $50.4$ & $13.3$ & $33.0$& $9.2$ \\
        Dola & $47.1$ & $13.8$ & $55.4$ & $15.7$ & $31.3$ & $9.0$\\
        VCD & $49.2$ & $14.8$ & $56.4$ & $15.5$ &  $32.8$& $8.9$\\
        OPERA & $45.4$ & $12.7$ & $46.2$ & $13.1$ & $34.0$& $8.7$\\
        SID & $44.2$ & $12.2$ & $44.8$ & $12.8$ & $32.8$ & $8.3$ \\
        \rowcolor[gray]{0.9} \eazy  & $\textbf{38.8}$ & $\textbf{11.4}$ & $\textbf{26.6}$ & $\textbf{8.9}$ & $\textbf{26.8}$& $\textbf{8.3}$\\
        \bottomrule
    \end{tabular}
    \caption{\textbf{Evaluation Results with CHAIR.} The lower, the better.}
    \label{tab:chair}
\end{table*}

\begin{table*}[h!]
\centering
\small
\begin{tabular}{llcc cc cc}
\toprule
\textbf{Method} & \multicolumn{2}{c}{\textbf{LLaVA-1.5}} & \multicolumn{2}{c}{\textbf{Shikra}} & \multicolumn{2}{c}{\textbf{LLaVA-NEXT}} \\
 \cmidrule(r){2-3} \cmidrule(l){4-5} \cmidrule(l){6-7}
 & Acc $\uparrow$ & F1 $\uparrow$ & Acc $\uparrow$ & F1 $\uparrow$ & Acc $\uparrow$ & F1 $\uparrow$ \\
\midrule
Greedy & $81.38$ & $82.20$ & $80.87$ & $81.05$ & $83.78$ & $82.24$ \\
Beam Search & $84.66$ & $84.60$ & $80.48$ & $81.81$ & $83.77$ & $81.69$ \\
Dola & $84.06$ & $84.62$ & $80.71$ & $81.19$ & $84.53$ & $84.77$ \\
VCD & $84.66$ & $84.52$ & $79.87$ & $81.06$ & $83.37$ & $83.95$ \\
OPERA & $84.88$ & $85.21$ & $79.99$ & $81.48$ & $84.37$ & $84.21$ \\
SID & $84.82$ & $85.50$ & ${80.02}$ & $81.23$ & $84.61$ & $85.32$ \\
\rowcolor[gray]{0.9} \eazy & $\textbf{84.97}$ & $\textbf{85.78}$ & $\textbf{80.96}$ & $\textbf{82.38}$ & $\textbf{84.91}$ & $\textbf{85.40}$ \\
\bottomrule
\end{tabular}
\caption{\textbf{Evaluation Result on POPE.} We take the average accuracy and F1 score of random, popular, and adversarial modes.}
\label{tab:pope}
\end{table*}

\subsection{Experiment Setting} 

\subsubsection{OH Mitigation Setting}

\noindent \textbf{Model.}We choose the LLaVA-1.5~\cite{liu2024improved}, Shikra~\cite{chen2023shikra} at 7B scale and LLaVA-Next~\cite{liu2024llavanext} at 8B scale for evaluation of OH mitigation. We set the maximum number of generated tokens as $512$ in our setting. Please refer to Appendix~\ref{sec:model_detail} for more model details.

\noindent \textbf{Baseline.} We compare our method \eazy with six \textit{training-free} methods including three LLM decoding strategies, greedy and beam search and Dola~\cite{chuang2024dola}, as well as three state-of-the-art hallucination mitigation methods for LVLM, VCD~\cite{leng2024mitigating}, OPERA~\cite{huang2024opera} and SID~\cite{huo2025selfintrospective}. We adopt greedy decoding for Dola, SID and EAZY. We set the beam number as $5$ and apply sampling for VCD and OPERA. We provide more implementation details in Appendix~\ref{sec:implementation}.

\subsubsection{OH Detection Setting}
\noindent\textbf{Baselines.} We chose two state-of-the-art \textit{training-free} OH detection methods as baselines to compare our method on Hall-COCO. Uncertainty (UT)~\cite{zhou2023analyzing} found that HOs often exhibit higher uncertainty scores during generation. Internal Confidence (IC)~\cite{jiang2024interpreting} found the maximum logit lens probability can be utilized for OH detection.

\noindent\textbf{Model and Benchmark.} We evaluate different OH detection methods on the LLaVA-1.5 7B model with greedy decoding. The maximum length of the generation is $512$ tokens. We use Hall-COCO as the benchmark for the open-end image description task. Each image has human-annotated real and hallucinatory objects. 

\noindent\textbf{Metric.} We follow~\cite{jiang2024interpreting} to treat the OH detection as a binary classification task, where real objects are the positive class and HOs as the negative. We use accuracy (Acc), precision (PR), recall, and F1 score as the evaluation metrics.


\begin{figure}
    \centering    \includegraphics[width=1\linewidth]{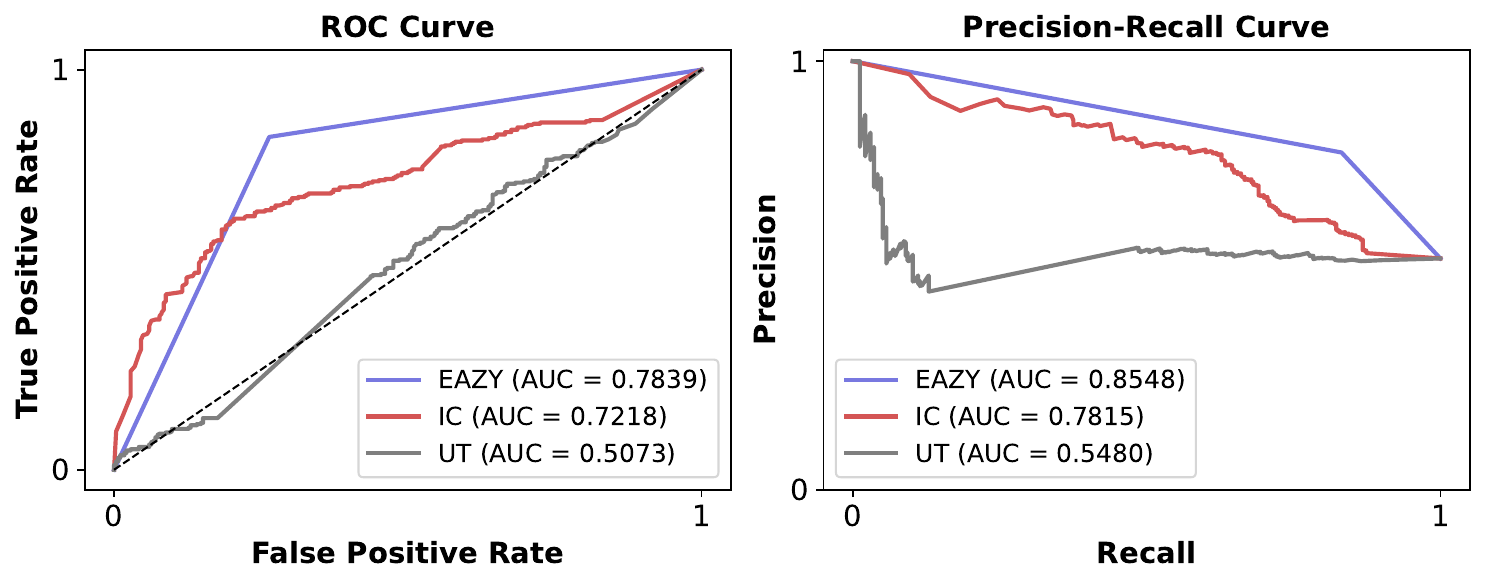}
    \caption{\textbf{Object Hallucination Detection Curves on Hall-COCO.} We present the Precision-Recall and ROC curves of the proposed OH detection method and baselines.}
    \label{fig:detection_curve}
\end{figure}

\begin{table}[!ht]
    \centering
    \begin{tabular}{cccc}
    \toprule
         \textbf{Metric} & \textbf{UT} & \textbf{IC} & \textbf{\eazy}  \\
         \midrule
         Acc & $50.57$ & $62.56$ & $\textbf{78.77}$  \\
         PR(RO) & $53.60$ & $61.93$ & $\textbf{78.41}$ \\
         PR(OH) & $42.77$ & $64.16$ & $\textbf{79.25}$\\
         Recall & $70.62$ & $81.60$ & $\textbf{83.38}$ \\
         F1 & $60.95$ & $70.42$ & $\textbf{80.82}$  \\
         \bottomrule
    \end{tabular}
    \caption{\textbf{OH Detection Results on Hall-COCO.} PR(RO) represents the precision of real objects (positive instances), while PR(OH) represents the precision of object hallucination (negative instances). \textbf{UT}~\cite{zhou2023analyzing} is the uncertainty detection method. \textbf{IC}~\cite{jiang2024interpreting} is the internal confidence method.}
    \label{tab:OH_detection}
\end{table}

\begin{table*}[!ht]
    \caption{MME \& MMBench result of LLaVA-1.5-7b}
    \centering
    \begin{tabular}{c|cccccc}
        \toprule
        Method  & Greedy & Dola & VCD & OPERA & SID & \eazy \\ \midrule
        MME     & 1510.7 & 1480.1 & 1488.7 & 1515.4 & 1520.2 &  1496.2\\
        MMBench & 64.3   & 63.8   & 63.9   & 64.4   & 65.1    &  64.1  \\
        \bottomrule
    \end{tabular}
    \label{tab:mme_mmb}
\end{table*}

\subsection{Evaluation}

\subsubsection{Evaluation Result on OH Mitigation}
We adopt CHAIR~\cite{rohrbach2018object} and POPE~\cite{li2023evaluating} as quantitative metrics to compare the hallucination mitigation performance between the proposed \eazy and baseline methods. We provide detailed descriptions of both metrics in Appendix~\ref{sec:eval_metric}. For the \textbf{CHAIR} result in Table~\ref{tab:chair}, we follow~\cite{huang2024opera,huo2025selfintrospective,gong2024damro,favero2024multi} to evaluate different LVLMs with $500$ randomly selected images from MSCOCO~\cite{lin2014microsoft} with the fixed prompt: \textit{``Please describe this image in detail."}. The proposed method \eazy surpasses the baselines by a large margin for three different LVLMs, validating the effectiveness of \eazy in open-end generation tasks. We show the average evaluation results on POPE in Table~\ref{tab:pope}, which comprises three datasets. \eazy attains the best average performance in random, popular, and adversarial sampling settings, outperforming all other methods. It can be observed that, under the POPE metric, the advantage of \eazy is not as pronounced as in the CHAIR results. We attribute this to the binary Yes-or-No question format of POPE, which may lead to generated content that does not include object tokens. Furthermore, the prompt \textit{``Is there an {object} in the image?"} already contains the object name, potentially impacting models' attention distribution.

\subsubsection{Evaluation Result on OH Detection}

We present the quantitative results of OH detection on LLaVA-1.5 in Table~\ref{tab:OH_detection} and Figure~\ref{fig:detection_curve}. We find that \eazy improves the accuracy $16.21\%$, $16.48\%$ in the precision of real objects, and $15.09\%$ in the precision of HOs, compared to the best baseline. In addition, \eazy improves the ROC AUC by $8.60\%$ and the PR AUC by $9.37\%$ over the baseline. This indicates the superior performance of \eazy on OH detection with a good balance of precision on both real and hallucinatory objects. We provide additional results for LLaVA-Next and Shikra models in Appendix~\ref{sec:extra_detection_exp}.


\subsection{Evaluation on MLLM benchmarks.} We evaluate \eazy on on two popular MLLM benchmark, MME~\cite{fu2024mmecomprehensiveevaluationbenchmark} and MMBench~\cite{liu2024mmbenchmultimodalmodelallaround}. Based on the result in Table~\ref{tab:mme_mmb}, we can see \eazy maintains the MLLM’s general capability on both benchmarks.

\subsection{Ablation Study}

\begin{figure}[h!]
    \centering
    \begin{minipage}{0.45\linewidth}
        \includegraphics[width=\textwidth]{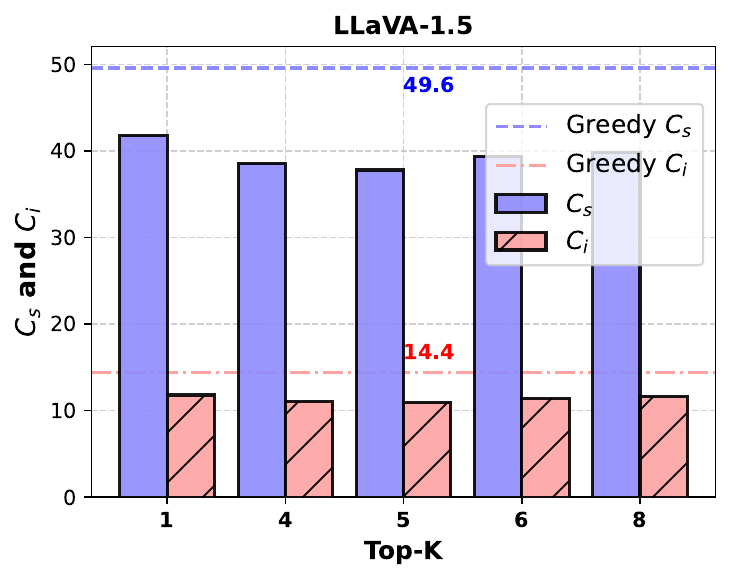} 
    \end{minipage}
    \begin{minipage}{0.45\linewidth}
        \centering
        \includegraphics[width=\textwidth]{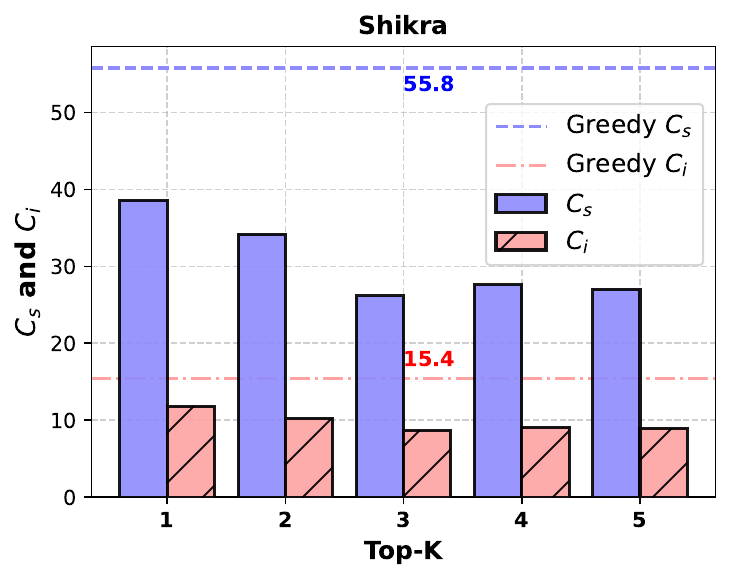}
    \end{minipage}
    \caption{CHAIR evaluation results as a function of $K$ on LLaVA-1.5 (left) and Shikra (right).} 
    \label{fig:topk_ablation} 
\end{figure}

\noindent\textbf{Impact of $K$ value selection. }
The choice of the value $K$ is critical to the mitigation performance of \eazy. In Figure~\ref{fig:topk_ablation}, we show the impact of different values $K$ for the LLaVA-1.5 and Shikra models on the CHAIR metric. We use the Greedy decoding performance as the reference line. It can be seen that LLaVA-1.5 achieves the best performance when $K=5$ and Shikra is on $K=3$.

\begin{table}[!h]
    \centering
    \begin{tabular}{c cc}
    \toprule
    \textbf{Position}  & $\text{CHAIR}_S \downarrow$ & $\text{CHAIR}_I \downarrow$ \\
    \midrule
      Input Token & $38.8$ & $11.4$\\
      ViT Output & $40.4$ &$11.8$ \\
      Image Patches  &$39.0$  &$11.4$  \\
    \bottomrule
    \end{tabular}
    \caption{Evaluation Result of Different Zero-Out Positions on CHAIR.}
    \label{tab:position_ablation}
\end{table}

\noindent \textbf{Impact of Zero-Out Position in the Pipeline. } In \eazy, we apply the zero-out of image tokens in the LLM input space, i.e., replace the target image tokens with zero embeddings. To transform the original image into image tokens, the LVLMs first utilize the ViT to process the image from image patches into patch embeddings, then use the linear projector or MLP to map the patch embeddings into image tokens. We apply the zero-out operation on three different positions in the image processing pipeline, including the input token space, the output space of ViT, and the original image patches. As shown in Table~\ref{tab:position_ablation}, applying the zero-out operation at different positions on the same set of image tokens (or patch embeddings) effectively mitigates object hallucination in the model response. This suggests that the visual biases introduced by hallucinatory image tokens originate directly from the image itself and are not strongly associated with modules such as ViT or Linear Projector.

\begin{table}[!h]
    \centering
    \begin{tabular}{c cc}
    \toprule
    \textbf{Methods}  & \textbf{Image w/o OH} & \textbf{Removed OH} \\
    \midrule
      \rowcolor[gray]{0.9} Zero-Out & $\textbf{68.27\%}$ & $\textbf{74.17\%}$ \\
      Average & $55.17\%$ & $66.67\%$\\
      Random & $58.56\%$ &$63.33\%$ \\
      Neighbor  &$55.17\%$  &$61.25\%$  \\
    \bottomrule
    \end{tabular}
    \caption{Evaluation Result of Top-K Image Token Processing Strategies on Hall-COCO.}
    \label{tab:processing_ablation}
\end{table}

\noindent \textbf{Impact of Image Token Processing Strategies. }
We explored how the different processing methods for the top-K image tokens impact the removal of object hallucination (OH). 
In Table~\ref{tab:processing_ablation}, we show the performance of four different processing strategies on the collected \textit{Hall-COCO} dataset when $K=5$. The zero-out method, which is used in \eazy, replaces the embeddings of the top-$K$ candidate HITs with zero embeddings. The average method replaces the top-K most attended image token embeddings with the average embedding of all the image tokens. The random method replaces the token embeddings of top-K candidate HITs with a randomly selected image token embedding. The neighbor method replaces the top-K candidate HITs embeddings with one of their neighbor image tokens adjunct to the target image token. Overall, our zero-out performs better in the percentage of images with fewer object hallucinations, images without object hallucinations, and removed object hallucinations compared to alternative strategies.

\section{Related Work}

\noindent \textbf{Cause of Object Hallucination.}
Hallucination~\cite{bai2024hallucination} affects the trustworthiness and reliability of LVLM for broad application. Specifically, object hallucination~\cite{rohrbach2018object}, which refers to describing HOs in the input images, is more harmful than other types of inaccurate description. Recent studies have proposed explaining the causes of hallucination from training data biases~\cite{zhou2023analyzing,wang2023llm,yu2024hallucidoctor} and language biases~\cite{leng2024mitigating,favero2024multi,huang2024opera,lee2023volcano}. In~\cite{liu2025paying,favero2024multi}, the authors found that LVLMs exhibit a preference for text tokens, causing the model to pay more attention on neighboring text tokens when generating new tokens.

\noindent \textbf{Mitigation of Object Hallucination.} Due to the complex structure and the large number of parameters in LVLMs, addressing hallucinations through retraining or fine-tuning is often prohibitively expensive and computationally intensive. As a result, many studies focus on mitigation during inference. A series of decoding strategy-based works~\cite{leng2024mitigating,woo2024don,wei2024dopra,favero2024multi} leverage contrastive decoding technique~\cite{li2022contrastive} to compare the model's original logit probability distribution with the logit output generated from processed images, thereby suppressing prior knowledge biases and statistical biases introduced by the training data. Meanwhile, methods such as OPERA~\cite{huang2024opera} and DAMRO~\cite{gong2024damro} have observed that sequence supervision-induced anchor tokens significantly increase the likelihood of hallucinations in subsequent tokens.  Compared to these works, we discovered that visual biases are the cause of most object hallucinations and developed a simple yet effective method by zeroing out the hallucinatory image tokens.

\section{Conclusion}
In this paper, we identify a new cause of object hallucination from the perspective of visual bias—Hallucinatory Image Tokens. A small number of image tokens with high attention scores are directly correlated with the presence of hallucinated objects in the generated text. Replacing these HITs with zero embeddings effectively removes hallucinated objects while minimally affecting real objects. Based on this finding, we propose \eazy, a training-free method for automatically detecting and mitigating object hallucinations. Through extensive experiments, we demonstrate the superior performance of \eazy in both hallucination detection and mitigation. We provide more discussion and visualization results in Appendix~\ref{sec:discussion} and~\ref{sec:app_case_study}.


\newpage

{
    \small
    \bibliographystyle{ieeenat_fullname}
    \bibliography{references}

\begin{thebibliography}{36}
\providecommand{\natexlab}[1]{#1}
\providecommand{\url}[1]{\texttt{#1}}
\expandafter\ifx\csname urlstyle\endcsname\relax
  \providecommand{\doi}[1]{doi: #1}\else
  \providecommand{\doi}{doi: \begingroup \urlstyle{rm}\Url}\fi

\bibitem[Bai et~al.(2024)Bai, Wang, Xiao, He, Han, Zhang, and Shou]{bai2024hallucination}
Zechen Bai, Pichao Wang, Tianjun Xiao, Tong He, Zongbo Han, Zheng Zhang, and Mike~Zheng Shou.
\newblock Hallucination of multimodal large language models: A survey.
\newblock \emph{arXiv preprint arXiv:2404.18930}, 2024.

\bibitem[Bird(2006)]{bird2006nltk}
Steven Bird.
\newblock Nltk: the natural language toolkit.
\newblock In \emph{Proceedings of the COLING/ACL 2006 Interactive Presentation Sessions}, pages 69--72, 2006.

\bibitem[Chen et~al.(2023)Chen, Zhang, Zeng, Zhang, Zhu, and Zhao]{chen2023shikra}
Keqin Chen, Zhao Zhang, Weili Zeng, Richong Zhang, Feng Zhu, and Rui Zhao.
\newblock Shikra: Unleashing multimodal llm's referential dialogue magic.
\newblock \emph{arXiv preprint arXiv:2306.15195}, 2023.

\bibitem[Chiang et~al.(2023)Chiang, Li, Lin, Sheng, Wu, Zhang, Zheng, Zhuang, Zhuang, Gonzalez, et~al.]{chiang2023vicuna}
Wei-Lin Chiang, Zhuohan Li, Zi Lin, Ying Sheng, Zhanghao Wu, Hao Zhang, Lianmin Zheng, Siyuan Zhuang, Yonghao Zhuang, Joseph~E Gonzalez, et~al.
\newblock Vicuna: An open-source chatbot impressing gpt-4 with 90\%* chatgpt quality.
\newblock \emph{See https://vicuna. lmsys. org (accessed 14 April 2023)}, 2\penalty0 (3):\penalty0 6, 2023.

\bibitem[Chuang et~al.(2024)Chuang, Xie, Luo, Kim, Glass, and He]{chuang2024dola}
Yung-Sung Chuang, Yujia Xie, Hongyin Luo, Yoon Kim, James~R. Glass, and Pengcheng He.
\newblock Dola: Decoding by contrasting layers improves factuality in large language models.
\newblock In \emph{The Twelfth International Conference on Learning Representations}, 2024.

\bibitem[Favero et~al.(2024)Favero, Zancato, Trager, Choudhary, Perera, Achille, Swaminathan, and Soatto]{favero2024multi}
Alessandro Favero, Luca Zancato, Matthew Trager, Siddharth Choudhary, Pramuditha Perera, Alessandro Achille, Ashwin Swaminathan, and Stefano Soatto.
\newblock Multi-modal hallucination control by visual information grounding.
\newblock In \emph{Proceedings of the IEEE/CVF Conference on Computer Vision and Pattern Recognition}, pages 14303--14312, 2024.

\bibitem[Fu et~al.(2024)Fu, Chen, Shen, Qin, Zhang, Lin, Yang, Zheng, Li, Sun, Wu, and Ji]{fu2024mmecomprehensiveevaluationbenchmark}
Chaoyou Fu, Peixian Chen, Yunhang Shen, Yulei Qin, Mengdan Zhang, Xu Lin, Jinrui Yang, Xiawu Zheng, Ke Li, Xing Sun, Yunsheng Wu, and Rongrong Ji.
\newblock Mme: A comprehensive evaluation benchmark for multimodal large language models, 2024.

\bibitem[Gong et~al.(2024)Gong, Ming, Wang, and Wei]{gong2024damro}
Xuan Gong, Tianshi Ming, Xinpeng Wang, and Zhihua Wei.
\newblock Damro: Dive into the attention mechanism of lvlm to reduce object hallucination.
\newblock \emph{arXiv preprint arXiv:2410.04514}, 2024.

\bibitem[Guan et~al.(2024)Guan, Liu, Wu, Xian, Li, Liu, Wang, Chen, Huang, Yacoob, et~al.]{guan2024hallusionbench}
Tianrui Guan, Fuxiao Liu, Xiyang Wu, Ruiqi Xian, Zongxia Li, Xiaoyu Liu, Xijun Wang, Lichang Chen, Furong Huang, Yaser Yacoob, et~al.
\newblock Hallusionbench: an advanced diagnostic suite for entangled language hallucination and visual illusion in large vision-language models.
\newblock In \emph{Proceedings of the IEEE/CVF Conference on Computer Vision and Pattern Recognition}, pages 14375--14385, 2024.

\bibitem[Huang et~al.(2024)Huang, Dong, Zhang, Wang, He, Wang, Lin, Zhang, and Yu]{huang2024opera}
Qidong Huang, Xiaoyi Dong, Pan Zhang, Bin Wang, Conghui He, Jiaqi Wang, Dahua Lin, Weiming Zhang, and Nenghai Yu.
\newblock Opera: Alleviating hallucination in multi-modal large language models via over-trust penalty and retrospection-allocation.
\newblock In \emph{Proceedings of the IEEE/CVF Conference on Computer Vision and Pattern Recognition}, pages 13418--13427, 2024.

\bibitem[Huo et~al.(2025)Huo, Xu, Zhang, Wang, Chen, and Zhao]{huo2025selfintrospective}
Fushuo Huo, Wenchao Xu, Zhong Zhang, Haozhao Wang, Zhicheng Chen, and Peilin Zhao.
\newblock Self-introspective decoding: Alleviating hallucinations for large vision-language models.
\newblock In \emph{The Thirteenth International Conference on Learning Representations}, 2025.

\bibitem[Jiang et~al.(2024)Jiang, Kachinthaya, Petryk, and Gandelsman]{jiang2024interpreting}
Nick Jiang, Anish Kachinthaya, Suzie Petryk, and Yossi Gandelsman.
\newblock Interpreting and editing vision-language representations to mitigate hallucinations.
\newblock \emph{arXiv preprint arXiv:2410.02762}, 2024.

\bibitem[Lee et~al.(2023)Lee, Park, Jo, and Seo]{lee2023volcano}
Seongyun Lee, Sue~Hyun Park, Yongrae Jo, and Minjoon Seo.
\newblock Volcano: mitigating multimodal hallucination through self-feedback guided revision.
\newblock \emph{arXiv preprint arXiv:2311.07362}, 2023.

\bibitem[Leng et~al.(2024)Leng, Zhang, Chen, Li, Lu, Miao, and Bing]{leng2024mitigating}
Sicong Leng, Hang Zhang, Guanzheng Chen, Xin Li, Shijian Lu, Chunyan Miao, and Lidong Bing.
\newblock Mitigating object hallucinations in large vision-language models through visual contrastive decoding.
\newblock In \emph{Proceedings of the IEEE/CVF Conference on Computer Vision and Pattern Recognition}, pages 13872--13882, 2024.

\bibitem[Li et~al.(2022)Li, Holtzman, Fried, Liang, Eisner, Hashimoto, Zettlemoyer, and Lewis]{li2022contrastive}
Xiang~Lisa Li, Ari Holtzman, Daniel Fried, Percy Liang, Jason Eisner, Tatsunori Hashimoto, Luke Zettlemoyer, and Mike Lewis.
\newblock Contrastive decoding: Open-ended text generation as optimization.
\newblock \emph{arXiv preprint arXiv:2210.15097}, 2022.

\bibitem[Li et~al.(2023)Li, Du, Zhou, Wang, Zhao, and Wen]{li2023evaluating}
Yifan Li, Yifan Du, Kun Zhou, Jinpeng Wang, Wayne~Xin Zhao, and Ji-Rong Wen.
\newblock Evaluating object hallucination in large vision-language models.
\newblock \emph{arXiv preprint arXiv:2305.10355}, 2023.

\bibitem[Lin et~al.(2014)Lin, Maire, Belongie, Hays, Perona, Ramanan, Doll{\'a}r, and Zitnick]{lin2014microsoft}
Tsung-Yi Lin, Michael Maire, Serge Belongie, James Hays, Pietro Perona, Deva Ramanan, Piotr Doll{\'a}r, and C~Lawrence Zitnick.
\newblock Microsoft coco: Common objects in context.
\newblock In \emph{Computer Vision--ECCV 2014: 13th European Conference, Zurich, Switzerland, September 6-12, 2014, Proceedings, Part V 13}, pages 740--755. Springer, 2014.

\bibitem[Liu et~al.(2024{\natexlab{a}})Liu, Li, Li, and Lee]{liu2024improved}
Haotian Liu, Chunyuan Li, Yuheng Li, and Yong~Jae Lee.
\newblock Improved baselines with visual instruction tuning.
\newblock In \emph{Proceedings of the IEEE/CVF Conference on Computer Vision and Pattern Recognition}, pages 26296--26306, 2024{\natexlab{a}}.

\bibitem[Liu et~al.(2024{\natexlab{b}})Liu, Li, Li, Li, Zhang, Shen, and Lee]{liu2024llavanext}
Haotian Liu, Chunyuan Li, Yuheng Li, Bo Li, Yuanhan Zhang, Sheng Shen, and Yong~Jae Lee.
\newblock Llava-next: Improved reasoning, ocr, and world knowledge, 2024{\natexlab{b}}.

\bibitem[Liu et~al.(2024{\natexlab{c}})Liu, Li, Wu, and Lee]{liu2024visual}
Haotian Liu, Chunyuan Li, Qingyang Wu, and Yong~Jae Lee.
\newblock Visual instruction tuning.
\newblock \emph{Advances in neural information processing systems}, 36, 2024{\natexlab{c}}.

\bibitem[Liu et~al.(2024{\natexlab{d}})Liu, Xue, Chen, Chen, Zhao, Wang, Hou, Li, and Peng]{liu2024survey}
Hanchao Liu, Wenyuan Xue, Yifei Chen, Dapeng Chen, Xiutian Zhao, Ke Wang, Liping Hou, Rongjun Li, and Wei Peng.
\newblock A survey on hallucination in large vision-language models.
\newblock \emph{arXiv preprint arXiv:2402.00253}, 2024{\natexlab{d}}.

\bibitem[Liu et~al.(2025)Liu, Zheng, and Chen]{liu2025paying}
Shi Liu, Kecheng Zheng, and Wei Chen.
\newblock Paying more attention to image: A training-free method for alleviating hallucination in lvlms.
\newblock In \emph{European Conference on Computer Vision}, pages 125--140. Springer, 2025.

\bibitem[Liu et~al.(2024{\natexlab{e}})Liu, Duan, Zhang, Li, Zhang, Zhao, Yuan, Wang, He, Liu, Chen, and Lin]{liu2024mmbenchmultimodalmodelallaround}
Yuan Liu, Haodong Duan, Yuanhan Zhang, Bo Li, Songyang Zhang, Wangbo Zhao, Yike Yuan, Jiaqi Wang, Conghui He, Ziwei Liu, Kai Chen, and Dahua Lin.
\newblock Mmbench: Is your multi-modal model an all-around player?, 2024{\natexlab{e}}.

\bibitem[Neo et~al.(2024)Neo, Ong, Torr, Geva, Krueger, and Barez]{neo2024towards}
Clement Neo, Luke Ong, Philip Torr, Mor Geva, David Krueger, and Fazl Barez.
\newblock Towards interpreting visual information processing in vision-language models.
\newblock \emph{arXiv preprint arXiv:2410.07149}, 2024.

\bibitem[nostalgebraist(2020)]{nostalgebraist2020logitlens}
nostalgebraist.
\newblock Interpreting gpt: The logit lens, 2020.
\newblock Accessed: 2024-12-21.

\bibitem[OpenAI and et~al.(2024)]{openai2024gpt4technicalreport}
OpenAI and Achiam et al.
\newblock Gpt-4 technical report, 2024.

\bibitem[Radford et~al.(2021)Radford, Kim, Hallacy, Ramesh, Goh, Agarwal, Sastry, Askell, Mishkin, Clark, et~al.]{radford2021learning}
Alec Radford, Jong~Wook Kim, Chris Hallacy, Aditya Ramesh, Gabriel Goh, Sandhini Agarwal, Girish Sastry, Amanda Askell, Pamela Mishkin, Jack Clark, et~al.
\newblock Learning transferable visual models from natural language supervision.
\newblock In \emph{International conference on machine learning}, pages 8748--8763. PMLR, 2021.

\bibitem[Rohrbach et~al.(2018)Rohrbach, Hendricks, Burns, Darrell, and Saenko]{rohrbach2018object}
Anna Rohrbach, Lisa~Anne Hendricks, Kaylee Burns, Trevor Darrell, and Kate Saenko.
\newblock Object hallucination in image captioning.
\newblock \emph{arXiv preprint arXiv:1809.02156}, 2018.

\bibitem[Sun et~al.(2023)Sun, Shen, Cao, Liu, Li, Shen, Gan, Gui, Wang, Yang, et~al.]{sun2023aligning}
Zhiqing Sun, Sheng Shen, Shengcao Cao, Haotian Liu, Chunyuan Li, Yikang Shen, Chuang Gan, Liang-Yan Gui, Yu-Xiong Wang, Yiming Yang, et~al.
\newblock Aligning large multimodal models with factually augmented rlhf.
\newblock \emph{arXiv preprint arXiv:2309.14525}, 2023.

\bibitem[Wang et~al.(2023{\natexlab{a}})Wang, Wang, Xu, Zhang, Gu, Jia, Yan, Zhang, and Sang]{wang2023llm}
Junyang Wang, Yuhang Wang, Guohai Xu, Jing Zhang, Yukai Gu, Haitao Jia, Ming Yan, Ji Zhang, and Jitao Sang.
\newblock An llm-free multi-dimensional benchmark for mllms hallucination evaluation.
\newblock \emph{arXiv preprint arXiv:2311.07397}, 2023{\natexlab{a}}.

\bibitem[Wang et~al.(2023{\natexlab{b}})Wang, Zhou, Xu, Shi, Zhao, Xu, Ye, Yan, Zhang, Zhu, et~al.]{wang2023evaluation}
Junyang Wang, Yiyang Zhou, Guohai Xu, Pengcheng Shi, Chenlin Zhao, Haiyang Xu, Qinghao Ye, Ming Yan, Ji Zhang, Jihua Zhu, et~al.
\newblock Evaluation and analysis of hallucination in large vision-language models.
\newblock \emph{arXiv preprint arXiv:2308.15126}, 2023{\natexlab{b}}.

\bibitem[Wei and Zhang(2024)]{wei2024dopra}
Jinfeng Wei and Xiaofeng Zhang.
\newblock Dopra: Decoding over-accumulation penalization and re-allocation in specific weighting layer.
\newblock In \emph{Proceedings of the 32nd ACM International Conference on Multimedia}, pages 7065--7074, 2024.

\bibitem[Woo et~al.(2024)Woo, Kim, Jang, Choi, and Kim]{woo2024don}
Sangmin Woo, Donguk Kim, Jaehyuk Jang, Yubin Choi, and Changick Kim.
\newblock Don't miss the forest for the trees: Attentional vision calibration for large vision language models.
\newblock \emph{arXiv preprint arXiv:2405.17820}, 2024.

\bibitem[Xiao et~al.(2023)Xiao, Tian, Chen, Han, and Lewis]{xiao2023efficient}
Guangxuan Xiao, Yuandong Tian, Beidi Chen, Song Han, and Mike Lewis.
\newblock Efficient streaming language models with attention sinks.
\newblock \emph{arXiv preprint arXiv:2309.17453}, 2023.

\bibitem[Yu et~al.(2024)Yu, Li, Wei, Pang, Ye, Qin, Tang, Tian, and Zhuang]{yu2024hallucidoctor}
Qifan Yu, Juncheng Li, Longhui Wei, Liang Pang, Wentao Ye, Bosheng Qin, Siliang Tang, Qi Tian, and Yueting Zhuang.
\newblock Hallucidoctor: Mitigating hallucinatory toxicity in visual instruction data.
\newblock In \emph{Proceedings of the IEEE/CVF Conference on Computer Vision and Pattern Recognition}, pages 12944--12953, 2024.

\bibitem[Zhou et~al.(2023)Zhou, Cui, Yoon, Zhang, Deng, Finn, Bansal, and Yao]{zhou2023analyzing}
Yiyang Zhou, Chenhang Cui, Jaehong Yoon, Linjun Zhang, Zhun Deng, Chelsea Finn, Mohit Bansal, and Huaxiu Yao.
\newblock Analyzing and mitigating object hallucination in large vision-language models.
\newblock \emph{arXiv preprint arXiv:2310.00754}, 2023.

\end{thebibliography}
}

\newpage
\clearpage
\newpage

\appendix

\section{Hall-COCO: Benchmarking for Object Hallucination Mitigation Evaluation}
\label{sec:hallcoco}

To validate that the phenomenon of HITs can be broadly applied to other images, we constructed a dataset, named \textit{Hall-COCO}, from MSCOCO~\cite{lin2014microsoft} where each image reliably induces object hallucination. 
The dataset creation involved several steps. First, we used GPT-4o~\cite{openai2024gpt4technicalreport} to detect hallucinations in the output of LLaVA. Next, human annotators identified specific object hallucinations in the model's responses. For each hallucinated object, we then identified the top $5$ image tokens with the highest attention scores. In cases where multiple hallucinated objects were present in the output, we applied a union operation to compile the set of relevant tokens. Through this process, we curated a dataset containing $200$ images that consistently induce object hallucinations.

\begin{table}[!h]
    \centering
    \begin{tabular}{ccc}
    \toprule
         \textbf{Image} &\textbf{Real Objects}  &  \textbf{OH} \\
         \midrule
         $200$ & $333$ &  $284$\\
    \bottomrule
    \end{tabular}
    \caption{The number of images, real objects and object hallucinations (OH) in Hall-COCO.}
    \label{tab:hall_COCO}
\end{table}

\section{Extended Experiment}
\label{sec:extended_exp}

\subsection{Model Details.}
\label{sec:model_detail}

\noindent \textbf{LLaVA-1.5.} The LLaVA-1.5 Model~\cite{liu2024visual} leverages the linear projector layer to align the vision and text modalities, with $576$ image tokens. It adopted the pre-trained vision transformer from CLIP\cite{radford2021learning} and the pre-trained language model as Vicuna\cite{chiang2023vicuna}.  

\noindent \textbf{Shikra.} The Shikra Model~\cite{chen2023shikra} introduces referential dialogue capabilities in multimodal large language models (MLLMs) by handling spatial coordinate inputs and outputs in natural language. It utilizes a vision encoder, an alignment layer, and a Vicuna-based language model without requiring extra vocabularies, position encoders, pre-/post-detection modules, or external plug-ins. The model enables interaction with images through natural pointing and location-aware responses, supporting tasks such as Referring Expression Comprehension (REC), PointQA, Image Captioning, and Visual Question Answering (VQA) with promising performance.

\noindent \textbf{LLaVA-Next.} The LLaVA-Next Model~\cite{liu2024llavanext} enhances multimodal capabilities by increasing the input image resolution up to $672\times 672$ pixels, supporting three aspect ratios. It utilizes an improved visual instruction tuning data mixture to bolster visual reasoning and OCR capabilities. The model employs a pre-trained vision transformer from CLIP and integrates with advanced language models like Vicuna and Mistral. 

\subsection{Evaluation Metric of OH Mitigation}
\label{sec:eval_metric}
\noindent \textbf{CHAIR } The Caption Hallucination Assessment with Image Relevance (CHAIR)~\cite{rohrbach2018object} is an evaluation tool designed to assess object hallucination issues in the image captioning task. CHAIR quantifies the degree of object hallucination from sentence-level,$\chair_S$, and image-level, $\chair_I$, which are calculated by the ratio of hallucinaty objects and the ground-truth label objects. Specifically, CHAIR is calculated as:
\begin{align}
    &\chair_S = \frac{|\{\text{hallucinated objects}\}|}{|\{\text{all mentioned objects}\}|}, \\
    &\chair_I = \frac{|\{\text{captions w/ hallucinated objects}\}|}{|\{\text{all captions}\}|}
\end{align}

\noindent \textbf{POPE } Polling-based Object Probing
Evaluation (POPE)~\cite{li2023evaluating} is a metric designed to evaluate object hallucination in LVLMs. During evaluation, POPE prompts LVLM with straightforward Yes-or-No questions about the presence of specific objects in an image (e.g., \texttt{"Is there an apple in the image?}") POPE offers three evaluation settings: \textit{Random, Popular, and Adversarial}. In Random Sampling, objects that do not exist in the image are selected randomly. In Popular Sampling, the top-$k$ most frequent objects in the entire image dataset that are absent from the current image are selected, where $k = [n/2]$. Lastly, Adversarial Sampling ranks objects by their co-occurrence frequency with ground-truth objects and selects the top-$k$ frequent ones that are not in the image. In our experiment, we show the average accuracy and F1 score result over three different sampling modes on the MSCOCO dataset~\cite{lin2014microsoft}.

\subsection{Implementation Detail}
\label{sec:implementation}

For LLaVA-1.5, we use the attention score from layer $15$ (index starting from $1$) to choose the top-$5$ most attended image tokens as the candidate HITs. For Shikre, we use the attention score from layer $18$ (index starting from $1$) to choose the top-$3$ most attended image tokens as the candidate HITs. For LLaVA-Next, we use the attention score from layer $15$ (index starting from $1$) to choose the top-$5$ most attended image tokens as the candidate HITs.

\subsection{Better OH Mitigation via Iterative Hallucinatory Objects Detection}

In \eazy, for each text object tokens, we take their top-$K$ most attended image tokens and take the union of all such sets as the zero-out list for HO detection inference. However, we realize that if we detect the hallucinatory objects iteratively, \eazy can achieve better hallucination mitigation performance. Specifically, assuming we extracted $n$ text object tokens from the initial generation, we may take each of such top-$K$ image tokens to zero out and observe if the corresponding object disappears from the new response. In such a case, candidate HITs from different text object tokens do not interfere with each other, enabling more effective detection and mitigation of hallucinations. We denote this iterative detection variant as \eazy(iterative) and represent its performance comparison with \eazy in Table~\ref{tab:iterative}. It can be observed that \eazy(iterative) generally improved the performance on CHAIR with MSCOCO dataset.

\begin{table*}[!htbp]
    \centering
    \begin{tabular}{l cc cc cc}
        \toprule
        \multirow{2}{*}{\textbf{Method}} & \multicolumn{2}{c}{\textbf{LLaVA-1.5}} & \multicolumn{2}{c}{\textbf{Shikra}} & \multicolumn{2}{c}{\textbf{LLaVA-NEXT}} \\
        \cmidrule(r){2-3} \cmidrule(l){4-5} \cmidrule(l){6-7}
        & $\text{CHAIR}_S \downarrow$ & $\text{CHAIR}_I \downarrow$ &  $\text{CHAIR}_S \downarrow$ & $\text{CHAIR}_I \downarrow$
        &  $\text{CHAIR}_S \downarrow$ & $\text{CHAIR}_I \downarrow$\\
        \midrule
        \eazy  & ${38.8}$ & ${11.4}$ & ${26.6}$ & ${8.9}$ & ${26.8}$& $\textbf{8.3}$\\
        \eazy (iterative) & $\textbf{37.8}$ & $\textbf{10.9}$ & $\textbf{26.2}$ & $\textbf{8.7}$ & $\textbf{25.6}$& ${8.5}$\\

        \bottomrule
    \end{tabular}
    \caption{\textbf{Evaluation Results with CHAIR between \eazy and \eazy (iterative).}}
    \label{tab:iterative}
\end{table*}

\subsection{Additional Evaluation Result of OH Detection}
\label{sec:extra_detection_exp}

We provide the OH detection evaluation result of LLaVA-Next and Shikra in this section. 

As shown in Tables~\ref{tab:OH_detection_llava_next} and ~\ref{tab:OH_detection_shikra}, applying the \eazy method to LLaVA-Next and Shikra for hallucinated object (HO) detection demonstrates that \eazy continues to achieve superior performance across the majority of key metrics. In particular, \eazy exhibits significant advantages in overall accuracy and the precision of hallucinated object detection. Compared to the best-performing baseline, \eazy improves accuracy by $7.65\%$ and $12.98\%$ on LLaVA-Next and Shikra, respectively, while achieving gains of $46.98\%$ and $23.56\%$ in PR(OH). This highlights its effectiveness in correctly identifying hallucinated objects. 

On the other hand, we observe that \eazy's performance in detecting real objects is less pronounced for both models. We attribute this to differences in how LLaVA-Next and Shikra process image-to-image token transformations compared to LLaVA-1.5. Specifically, Shikra maps $576$ image patch embeddings into $256$ image tokens through an MLP, which causes visual information to intertwine and interfere with one another. Consequently, zeroing out specific image tokens may introduce inaccuracies. For LLaVA-Next, the model employs the AnyRes technique to split and resize the original image, requiring the removal of more image tokens than LLaVA-1.5. This likely impacts the detection of real objects by reducing the amount of preserved visual information.

Nevertheless, \eazy consistently demonstrates strong performance in object hallucination (OH) detection across both LLaVA-Next and Shikra, achieving the highest F1 scores of $80.12\%$ and $76.61\%$, respectively. This indicates its superior ability to balance precision and recall, making it the most effective method overall. Particularly, \eazy achieves the highest recall on both models ($92.67\%$ on LLaVA-Next and $84.92\%$ on Shikra), significantly surpassing the baseline methods. This suggests that \eazy effectively minimizes false negatives, ensuring that a larger proportion of hallucinated objects are correctly identified.

\begin{figure}
    \centering
    \includegraphics[width=1\linewidth]{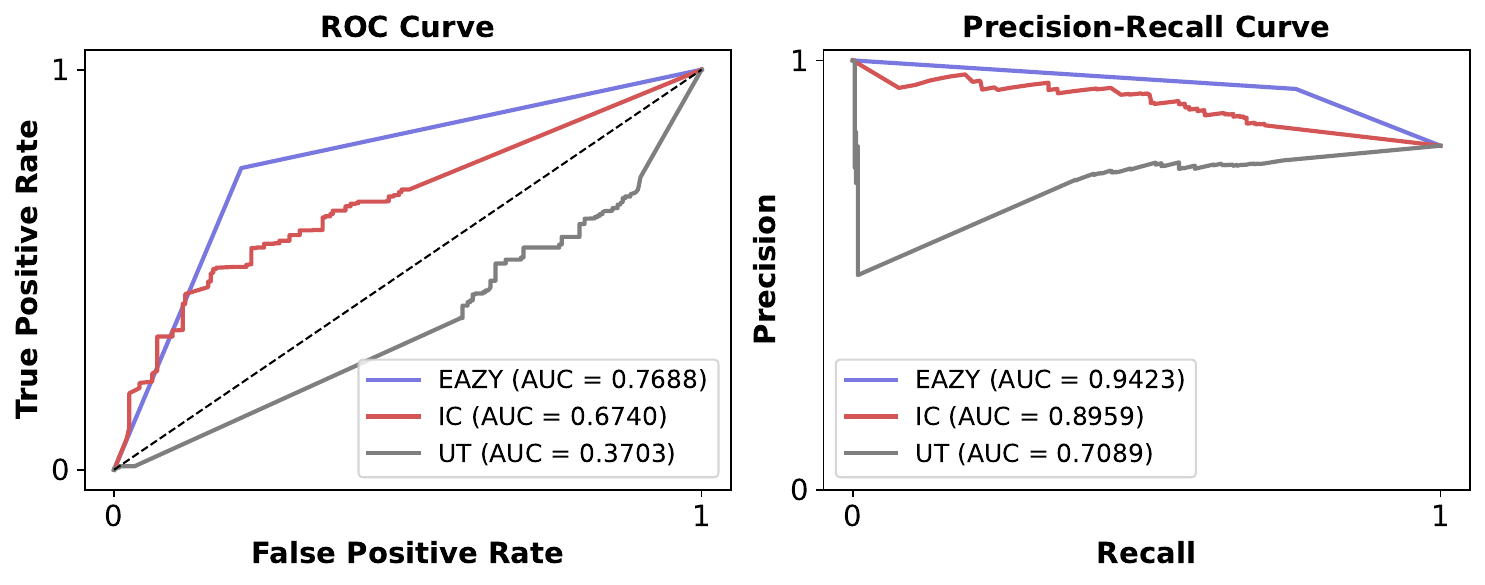}
    \caption{\textbf{Object Hallucination Detection Curves of LLaVA-Next on Hall-COCO.} We present the Precision-Recall and ROC curves of the proposed OH detection method and baselines.}
    \label{fig:detection_curve_llava_next}
\end{figure}

\begin{figure}
    \centering
    \includegraphics[width=1\linewidth]{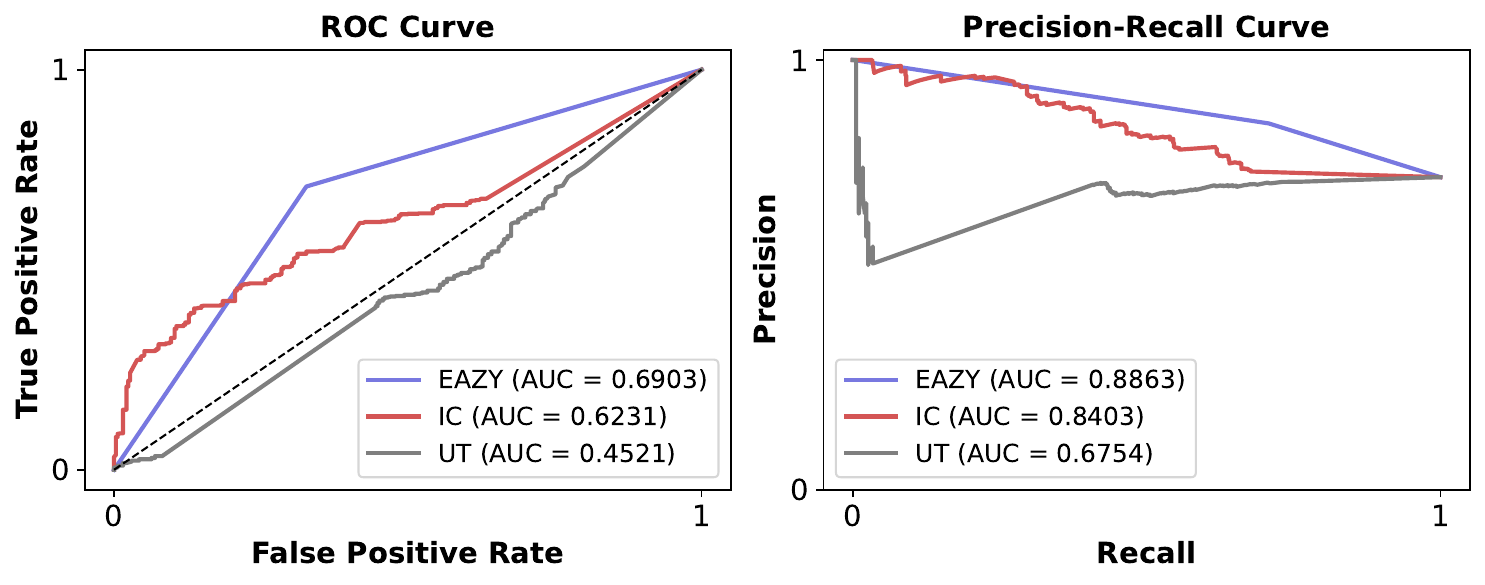}
    \caption{\textbf{Object Hallucination Detection Curves of Shikra on Hall-COCO.} We present the Precision-Recall and ROC curves of the proposed OH detection method and baselines.}
    \label{fig:detection_curve_shikra}
\end{figure}

Furthermore, \eazy achieves the highest AUC scores in both ROC and Precision-Recall (PR) curve evaluations as shown in Figure~\ref{fig:detection_curve_llava_next} and ~\ref{fig:detection_curve_shikra}. On LLaVA-Next, \eazy attains an AUC of $0.7688$ for ROC and $0.9423$ for PR, while on Shikra, it achieves $0.6903$ for ROC and $0.8863$ for PR. These results highlight \eazy’s ability to maintain a favorable trade-off between precision and recall, reinforcing its robustness across different vision-language models.

\begin{table}[!ht]
    \centering
    \begin{tabular}{cccc}
    \toprule
         \textbf{Metric} & \textbf{UT} & \textbf{IC} & \textbf{\eazy}  \\
         \midrule
         Acc & $49.05$ & $64.29$ & $\textbf{71.94}$  \\
         PR(RO) & $\textbf{87.75}$ & $87.39$ & ${70.57}$ \\
         PR(OH) & $24.68$ & $30.51$ & $\textbf{77.49}$\\
         Recall & $42.32$ & $64.77$ & $\textbf{92.67}$ \\
         F1 & $57.10$ & $74.40$ & $\textbf{80.12}$  \\
         \bottomrule
    \end{tabular}
    \caption{\textbf{OH Detection Results using LLaVA-Next on Hall-COCO.} PR(RO) represents the precision of real objects (positive instances), while PR(OH) represents the precision of object hallucination (negative instances). \textbf{UT}~\cite{zhou2023analyzing} is the uncertainty detection method. \textbf{IC}~\cite{jiang2024interpreting} is the internal confidence method.}
    \label{tab:OH_detection_llava_next}
\end{table}

\begin{table}[!ht]
    \centering
    \begin{tabular}{cccc}
    \toprule
         \textbf{Metric} & \textbf{UT} & \textbf{IC} & \textbf{\eazy}  \\
         \midrule
         Acc & $49.59$ & $56.03$ & $\textbf{69.01}$  \\
         PR(RO) & $77.33$ & $\textbf{81.87}$ & ${69.77}$ \\
         PR(OH) & $43.41$ & $34.79$ & $\textbf{66.97}$\\
         Recall & $55.60$ & $50.80$ & $\textbf{84.92}$ \\
         F1 & $30.45$ & $62.69$ & $\textbf{76.61}$  \\
         \bottomrule
    \end{tabular}
    \caption{\textbf{OH Detection Results using Shikra on Hall-COCO.} PR(RO) represents the precision of real objects (positive instances), while PR(OH) represents the precision of object hallucination (negative instances). \textbf{UT}~\cite{zhou2023analyzing} is the uncertainty detection method. \textbf{IC}~\cite{jiang2024interpreting} is the internal confidence method.}
    \label{tab:OH_detection_shikra}
\end{table}

\section{Efficiency Analysis}
\begin{figure}[!h]
    \centering
    \includegraphics[width=0.7\linewidth]{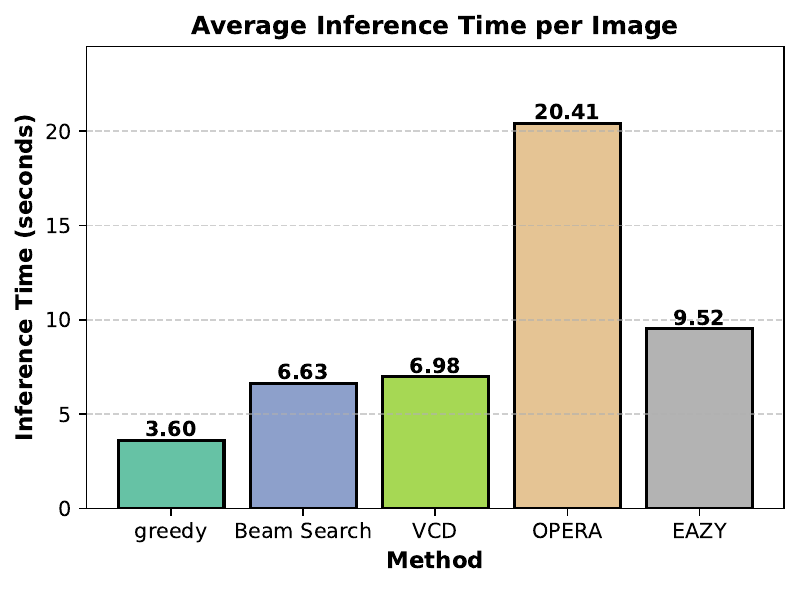}
    \caption{The average inference time of different methods on CHAIR evaluation.}
    \label{fig:infer_time}
\end{figure}

We evaluate the inference time (in seconds) on the LLaVA-1.5 7B under the CHAIR evaluation setting, which is close to the real-world application scenario. As shown in Figure~\ref{fig:infer_time}, we present the average inference time over $500$ image with the prompt "Please describe the image in detail.". We evaluate the different methods on NVIDIA A6000 GPUs. It can be observed the proposed \eazy method has $50\%$ lower inference time than OPERA, while it brings significant performance improvement on object hallucination mitigation.

\section{Statistic on Image Token Zeroing-out}

We measured the average number of image tokens zeroed out when applying \eazy across different LVLMs, as well as the ratio of this number to the total image tokens. As shown in Table~\ref{tab:token_remove_ratio}, LLaVA-1.5 zeroes out an average of $8.98$ image tokens out of $576$, resulting in a ratio of $1.56\%$. Shikra removes $5.23$ image tokens on average from a total of $256$, yielding a slightly higher ratio of $2.04\%$. LLaVA-Next, which processes significantly more image tokens ($2880$ in total), removes an average of $40.95$ tokens, with a ratio of $1.42\%$. These results demonstrate that \eazy requires modifying only a small fraction of image tokens to effectively mitigate hallucinations, highlighting its efficiency and minimal impact on the overall visual representation.

\begin{table}[!ht]
    \centering
    \small
    \begin{tabular}{cccc}
    \toprule
         \textbf{Model} & \textbf{LLaVA-1.5} & \textbf{Shikra} & \textbf{LLaVA-Next}  \\
         \midrule
         \# Average Zeroed-out & $8.98$ & $5.23$ & $40.95$  \\
         \# Total Image Token & $576$ & $256$ & $2880$ \\
         Ratio($\%$) & $1.56$ & $2.04$ & $1.42$\\
         \bottomrule
    \end{tabular}
    \caption{The statistic result of the average zeroed-out image token numbers by \eazy, the total number of image tokens and the ratio between them for each model. }
    \label{tab:token_remove_ratio}
\end{table}

\section{Additional Ablation Study}

\subsubsection{Impact of Zero-out Layer Choice}
We evaluate the impact of different zero-out layer choices for hallucination mitigation. As shown in Table~\ref{tab:chair_layer}, we compare the CHAIR scores when zero-out is applied on layer $10$, $15$, and $25$, respectively. It can be seen that layer $15$ obtained the best performance, which aligned with our analysis in Section~\ref{sec:localize}.

\begin{table}[]
\vspace{-4.5mm}
\caption{\eazy on different layers of LLaVA-1.5-7b}
    \centering
    \begin{tabular}{c|cc}
        \toprule
        Layer & $\text{CHAIR}_S \downarrow$ & $\text{CHAIR}_I \downarrow$ \\ \midrule
        10  & 39.8 & \textbf{11.3} \\
        15 & \textbf{38.8}   & {11.4}  \\
        25 & 41.7 & 11.9 \\
        \bottomrule
    \end{tabular}
    \label{tab:chair_layer}
\end{table}

\section{Visual Bias is the Main Cause of Object Hallucination.}

The causes of Object Hallucination (OH) are complex and may include factors such as training data, model architecture, cross-modal alignment, training paradigms, and inference methods. We focus on factors that may lead to OH during the inference process, further exploring the impact of visual bias on OH and its proportion among different causes.

To simplify the problem, we categorize the factors causing OH during inference into three main categories:

\begin{itemize}
    \item \textbf{Visual Bias}
    \begin{itemize}
        \item {Image Ambiguity}: Low-resolution, misleading attributes (color, texture, shape), shadow, etc.
        \item {Modality Alignment}: The gap between the mapped visual features and text features
    \end{itemize}
    \item \textbf{Language Prior}: 
    \begin{itemize}
        \item Statistical learning bias of the language model leading to hallucination, including prior knowledge, grammar, co-occurrence, etc.
    \end{itemize}
    \item \textbf{Others:}
    \begin{itemize}
        \item Training Data Distribution Bias, Distribution imbalance hallucination, etc.
    \end{itemize}
\end{itemize}

For a hallucinatory object, its emergence may result from multiple overlapping causes. If a specific factor is altered, causing the hallucinated object (HO) to disappear or change, we consider that factor to be the dominant cause of the HO. 

Following with~\cite{leng2024mitigating}, we gradually add Gaussian noise on the image input to erase and weaken visual information. In such a case, we reduce the impact of visual bias while amplifying the influence of language priors and other factors. To evaluate the impact of visual bias in image inputs on object hallucination generation, we conducted experiments using LLaVA-1.5 on the Hall-COCO dataset. Specifically, we analyzed how the generation of hallucinated objects (HOs) changes when different levels of noise are added to the images. Specifically, we define New Hallucination as a newly generated hallucinatory object that appears after adding noise to the image. We define Stable Hallucination as a previously known hallucinatory object that persists even after applying noise. This suggests that Stable Hallucinations rely less on visual input and are instead more influenced by language priors.

\begin{figure}
    \centering
    \includegraphics[width=1\linewidth]{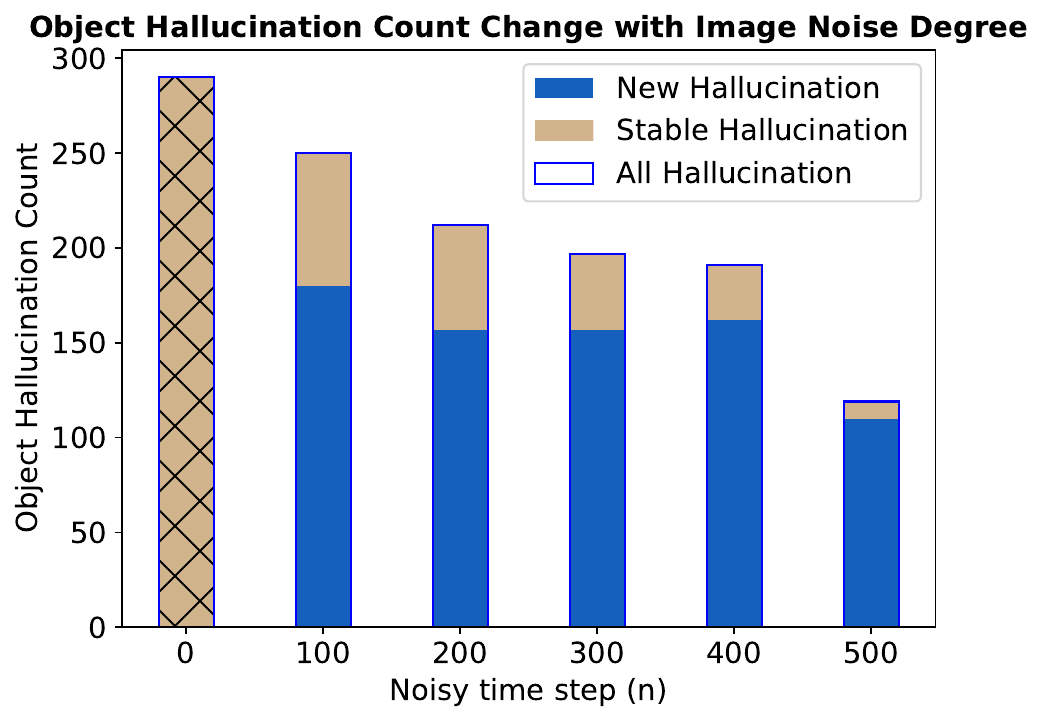}
    \caption{The amount change of different object hallucination along with the noise steps.}
    \label{fig:count_noise}
\end{figure}

As shown in Figure~\ref{fig:count_noise}, we observe that as the noise time step increases, the image gradually becomes indistinguishable from pure Gaussian noise. First, we note that the number of hallucinatory objects gradually decreases. This is primarily due to the reduction in visual information, which leads the model to generate shorter text responses. 

On the other hand, we find that the proportion of New Hallucinations significantly exceeds that of Stable Hallucinations, and the proportion of Stable Hallucinations continues to decline. After simply adding noise for $100$ steps, the vast majority of hallucinatory objects are New Hallucinations, indicating that visual bias is a key factor driving the variation and generation of object hallucinations.

\begin{figure}
    \centering
    \includegraphics[width=0.9\linewidth]{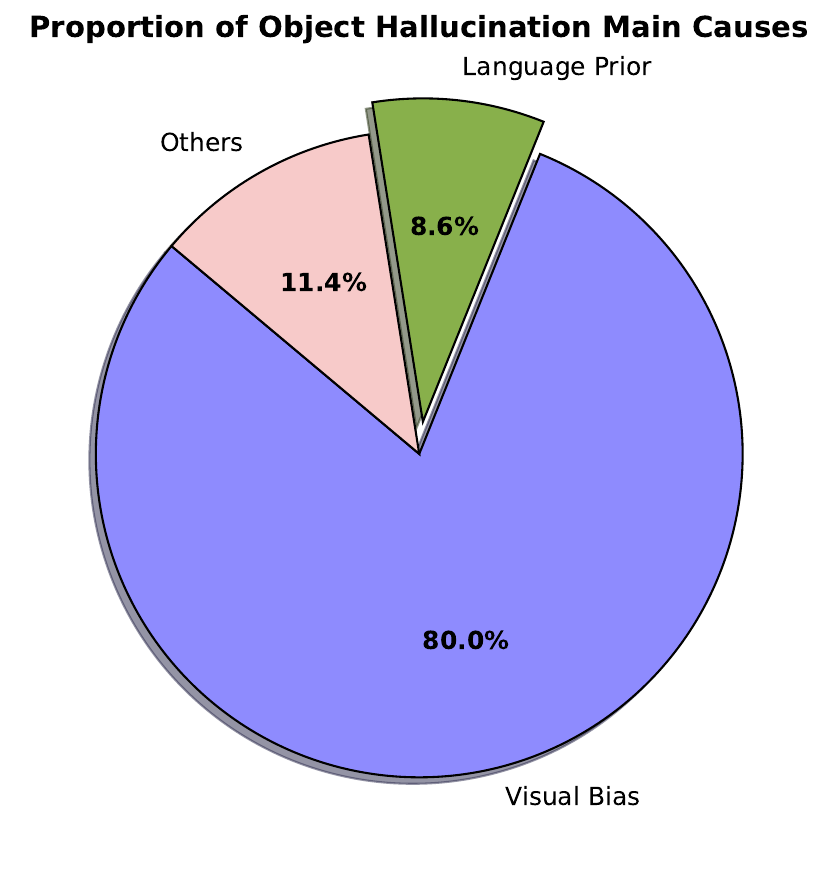}
    \caption{The proportion of different OH causes. Visual bias caused $80\%$ of object hallucination changes.}
    \label{fig:cause_proportion}
\end{figure}

To further estimate the relative impact of visual bias and language prior on object hallucination (OH), we fix the noise time step at 300. This setting largely removes visual information while avoiding excessive noise that would render the model’s output meaningless. We track the intersection of hallucinatory objects (HOs) that persist after applying \eazy on the original image and those that remain after adding 300-step Gaussian noise. Our findings indicate that 20\% of all HOs are Stable Hallucinations, among which 43\% cannot be eliminated by \eazy. As shown in Figure~\ref{fig:cause_proportion}, this suggests that visual bias accounts for $80\%$ of OH variations, while $8.6\%$ of OHs are primarily influenced by language prior. The remaining $11.4\%$ can be attributed to other factors.

\section{Hallucinatory objects are looking at image regions visually close to them.}
\label{sec:real_HO_heatmap}
\begin{figure}[!h]
    \centering
    \includegraphics[width=1\linewidth]{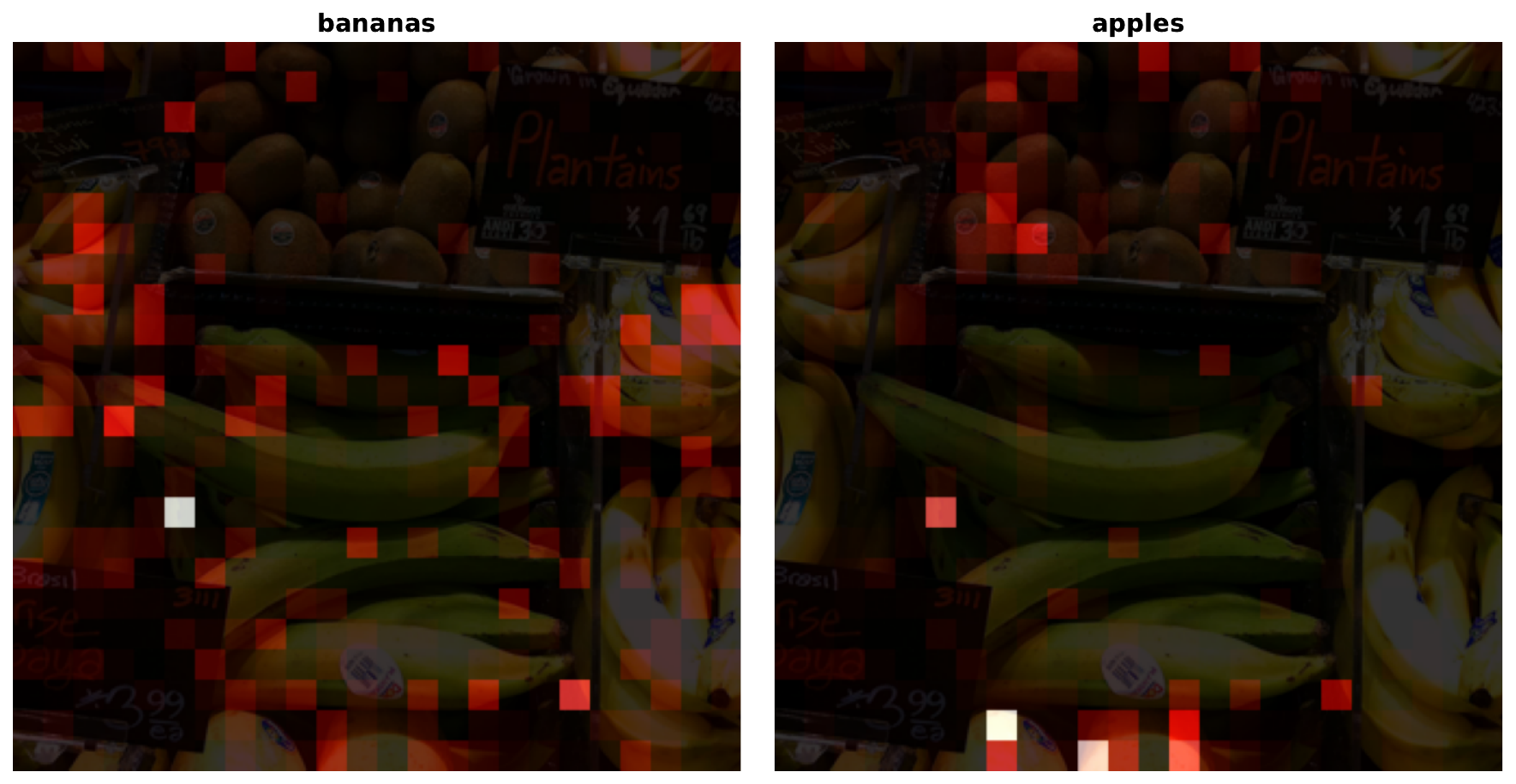}
    \caption{\textbf{Left:} The attention heatmap of the real object``\textit{bananas}"; \textbf{Right:} The attention heatmap of the hallucinatory object``\textit{apples}".}
    \label{fig:fruit_heatmap}
\end{figure}

\begin{figure}[!h]
    \centering
    \includegraphics[width=1\linewidth]{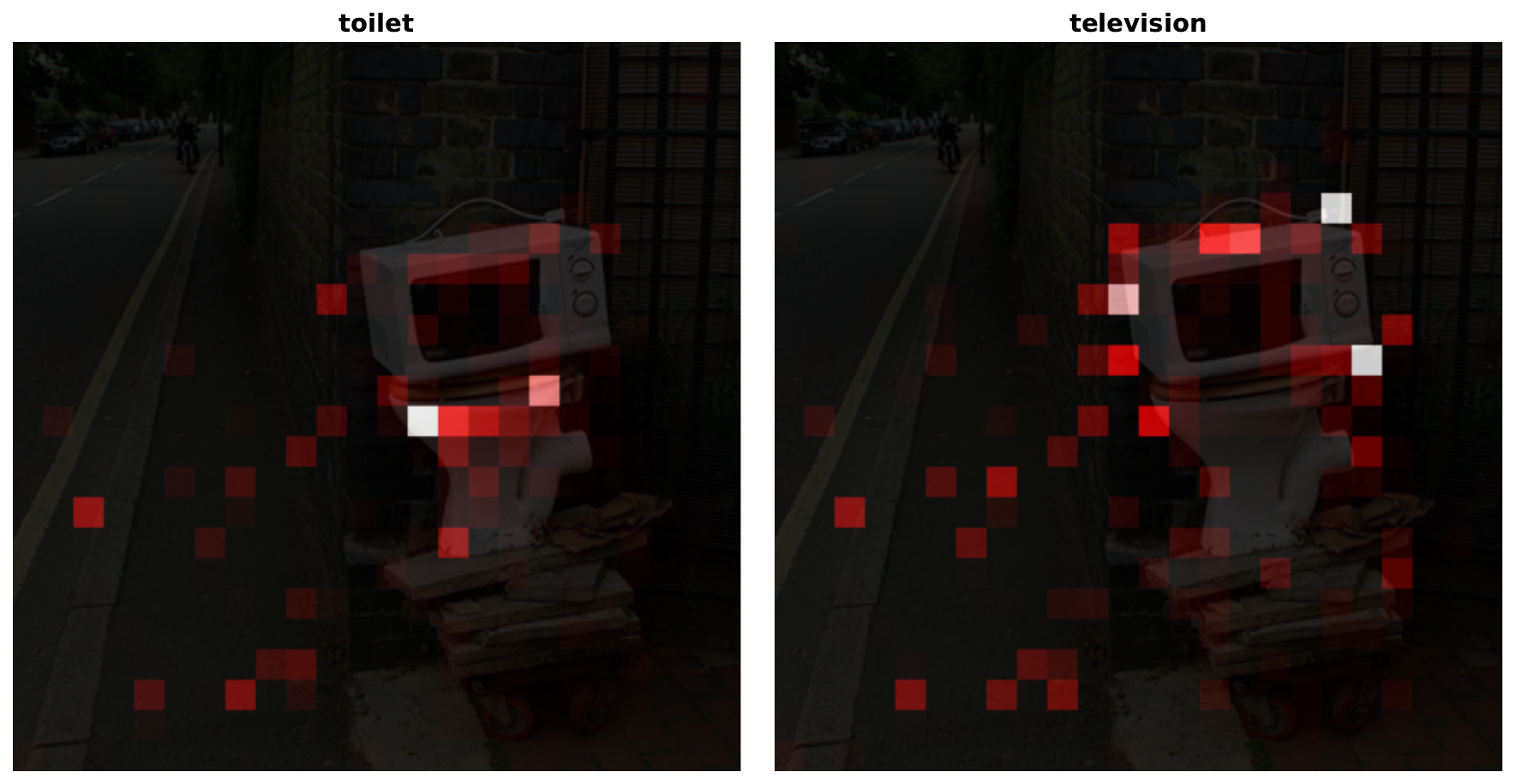}
    \caption{\textbf{Left:} The attention heatmap of the real object``\textit{toilet}"; \textbf{Right:} The attention heatmap of the hallucinatory object``\textit{television}".}
    \label{fig:toilet_heatmap}
\end{figure}

\begin{figure}[!h]
    \centering
    \includegraphics[width=1\linewidth]{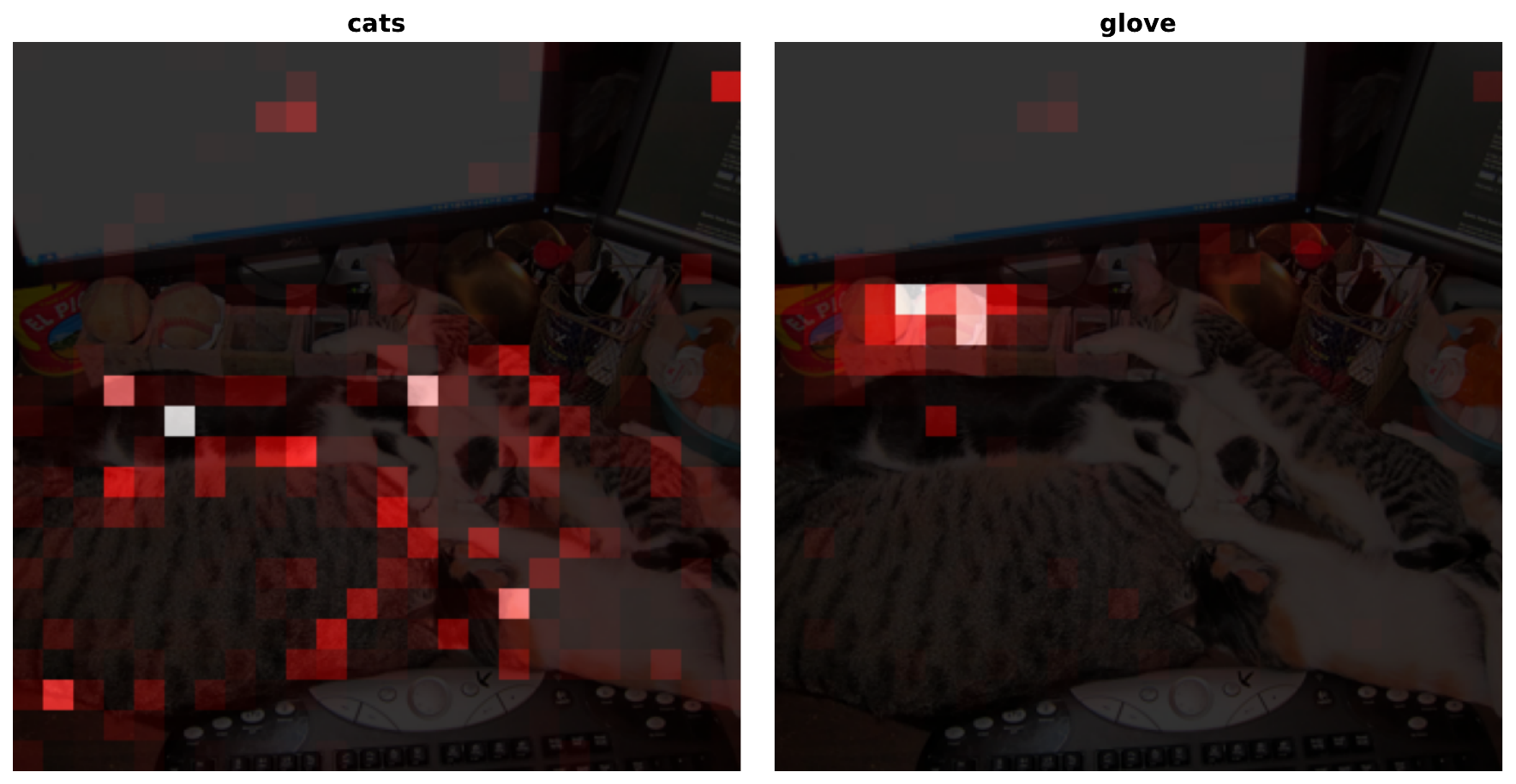}
    \caption{\textbf{Left:} The attention heatmap of the real object``\textit{cats}"; \textbf{Right:} The attention heatmap of the hallucinatory object``\textit{glove}".}
    \label{fig:cat_heatmap}
\end{figure}

\begin{figure}[!h]
    \centering
    \includegraphics[width=1\linewidth]{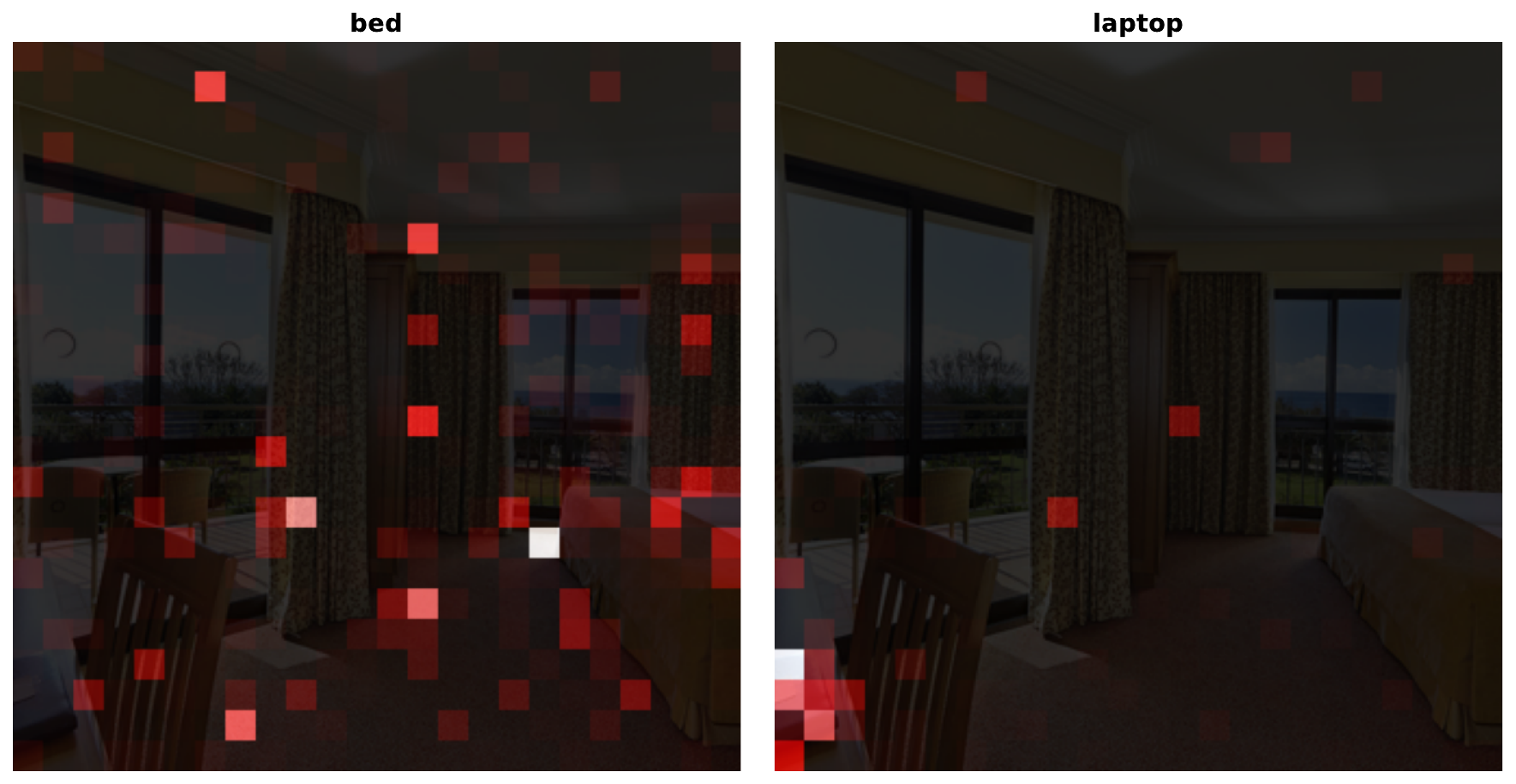}
    \caption{\textbf{Left:} The attention heatmap of the real object``\textit{bed}"; \textbf{Right:} The attention heatmap of the hallucinatory object``\textit{laptop}".}
    \label{fig:bed_heatmap}
\end{figure}

In this section, we compare the attention score distribution of real objects and hallucinatory objects over image tokens. We visualize the text-to-image token-wise attention heatmap using the attention from the 15th layer of the LLaVA-1.5 model. As shown in Figure~\ref{fig:fruit_heatmap}, ~\ref{fig:cat_heatmap}, ~\ref{fig:bed_heatmap} and ~\ref{fig:toilet_heatmap}, the attention distribution for real objects is typically concentrated on their corresponding image anchors. In contrast, hallucinatory objects follow two distinct attention patterns: One, a dispersed attention distribution across different image regions; and two, a highly concentrated focus on specific image regions that induces visual bias. Despite these differences, both patterns share a common characteristic - attended regions often have some visual resemblance to the hallucinated object.

These regions often exhibit visual similarities to the hallucinated objects. For example, in Figure~\ref{fig:fruit_heatmap}, the attention for \textit{apple} is distributed over the \textit{papaya} and \textit{kiwi} regions— \textit{papaya} shares a similar texture and color, while \textit{kiwi} has a comparable shape. In Figure~\ref{fig:toilet_heatmap}, the hallucinated object \textit{television} primarily attends to the \textit{microwave}, likely due to the visual resemblance of its glass panel and white frame to a television. In Figure~\ref{fig:cat_heatmap}, the hallucinated \textit{baseball glove} directs all its attention to the baseball itself. Similarly, in Figure~\ref{fig:bed_heatmap}, the hallucinated \textit{laptop} focuses on a black square notebook on the desk, which has a shape similar to a laptop.

\section{Failure Case Analysis.}
\label{sec:failure_Case}
\begin{figure*}[!h]
    \centering
    \includegraphics[width=0.9\linewidth]{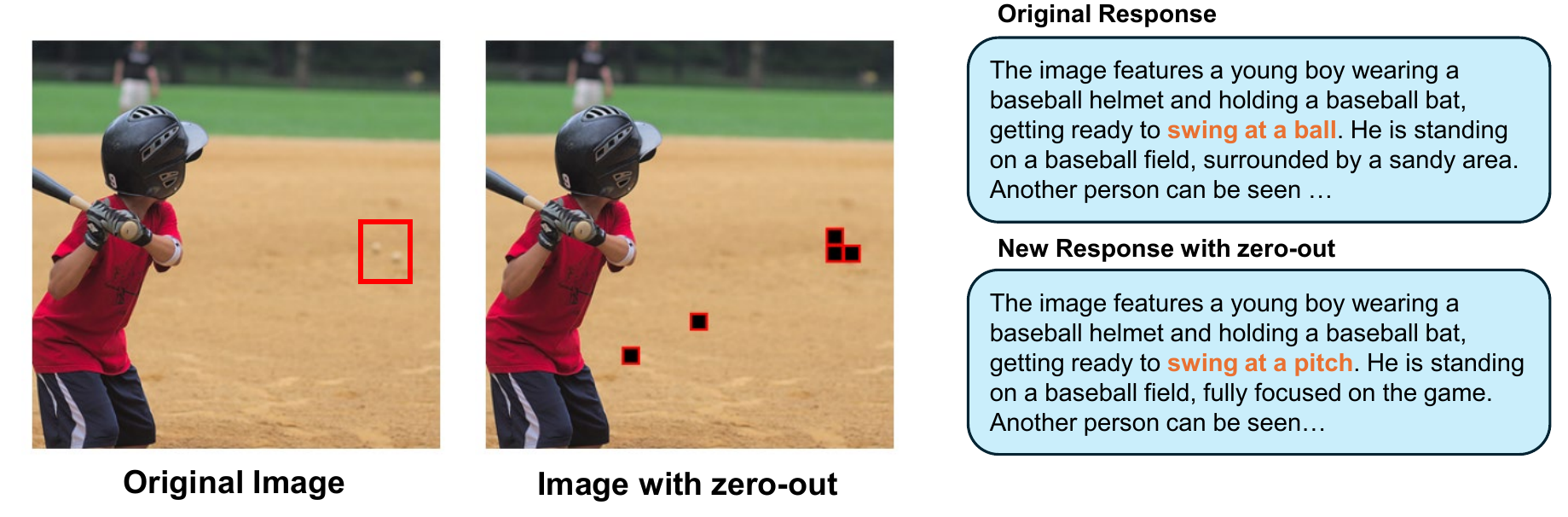}
    \caption{\textbf{Failure case of object ``ball".} The original response successfully recognizes the ``ball" on the ground. Due to the small size, applying zero-out directly masks the ball object from the image, resulting in the disappearance of the ``ball" in the new response}
    \label{fig:baseball_fail}
\end{figure*}

\begin{figure*}[!h]
    \centering
    \includegraphics[width=0.9\linewidth]{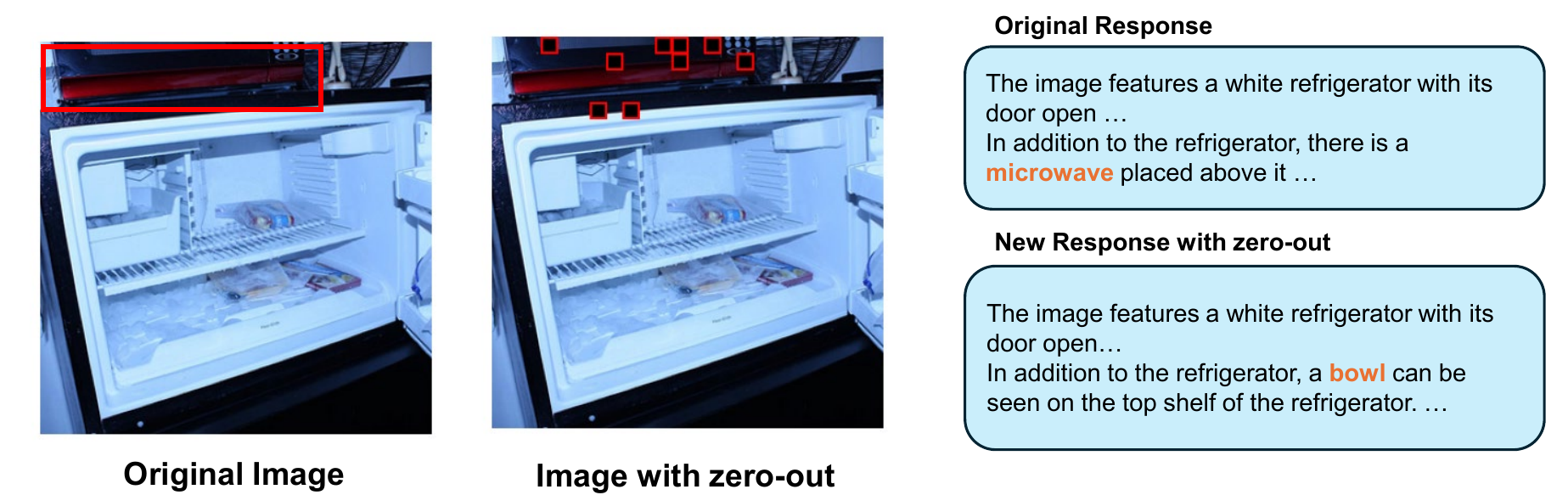}
    \caption{\textbf{Failure case of object ``microwave".} The word ``microwave" is split into ``mic", ``row" and ``ave", which significantly increase the number of image tokens for zeroing out. In addition, this object is located on the edge of the image, with only part of it shown in the image. The new response does not have the ``microwave" in it.}
    \label{fig:ref_fail}
\end{figure*}

We identify two typical cases of real objects being removed after applying \eazy. As shown in Figure~\ref{fig:baseball_fail}, the small-size objects have the chance to be entirely removed from the image input, causing the model unable to recognize the objects. 
When the objects are located at the edge of the image with only part of them revealed, additionally, longer words are tokenized into multiple tokens, it may lead to the excessive removal of image tokens. Such an example can be seen in Figure~\ref{fig:ref_fail}.

\section{Discussion and Future Work}
\label{sec:discussion}

\subsection{Summary of Key Findings}
Our work provides a new perspective on understanding the causes and mechanisms of object hallucination in LVLMs—visual bias. We first demonstrate that LVLMs identify and extract object-related information from image tokens in their early-middle to late-middle layers. This finding leads us to an important and intriguing phenomenon—\textbf{Hallucinatory Image Tokens} (HITs). By removing only a small subset of these tokens, we can significantly mitigate object hallucination in LVLM-generated outputs. 

Building on this discovery, we propose \eazy, a training-free approach that effectively detects and mitigates OH. By replacing the top-$K$ candidate HITs with zero embeddings, \eazy successfully suppresses hallucinated objects while preserving real object information. Our experiments demonstrated the superior performance of \eazy on HO detection and mitigation.

\subsection{Limitations and Future Work}

Our work leads to two promising questions. The first is achieving a deeper understanding of the HITs phenomenon. We found that existing interpretability tools, such as the Logit Lens, are insufficient for fully explaining and identifying all HITs. In future work, we aim to gain a more comprehensive understanding of HITs and leverage advanced interpretability tools to better distinguish them from their top-$K$ candidates, thereby significantly improving the efficiency of \eazy in zeroing out image tokens.

The second lies in \eazy’s handling of real objects. As discussed in Appendix~\ref{sec:failure_Case}, \eazy may occasionally misclassify real objects—particularly those that are small in size, tokenized into multiple tokens, or located at the image boundaries—and erroneously remove them from the generated response. To address this, we plan to explore the following directions:
\begin{enumerate}
    \item Performing multiple inferences to assess the robustness of different object tokens.
    \item Incorporating additional features, such as uncertainty estimation, to improve the detection accuracy.
    \item Integrating external vision models to refine object identification and mitigate errors.
\end{enumerate}

We hope that the findings of this work will contribute to a deeper understanding of object hallucination in LVLMs, inspire future research, and enable broader applications of the HITs phenomenon in vision-language models.

\section{Case Study}
Please refer to the next page for more visualization cases.

\label{sec:app_case_study}

\begin{figure*}[!h]
    \centering
    \includegraphics[width=0.9\linewidth]{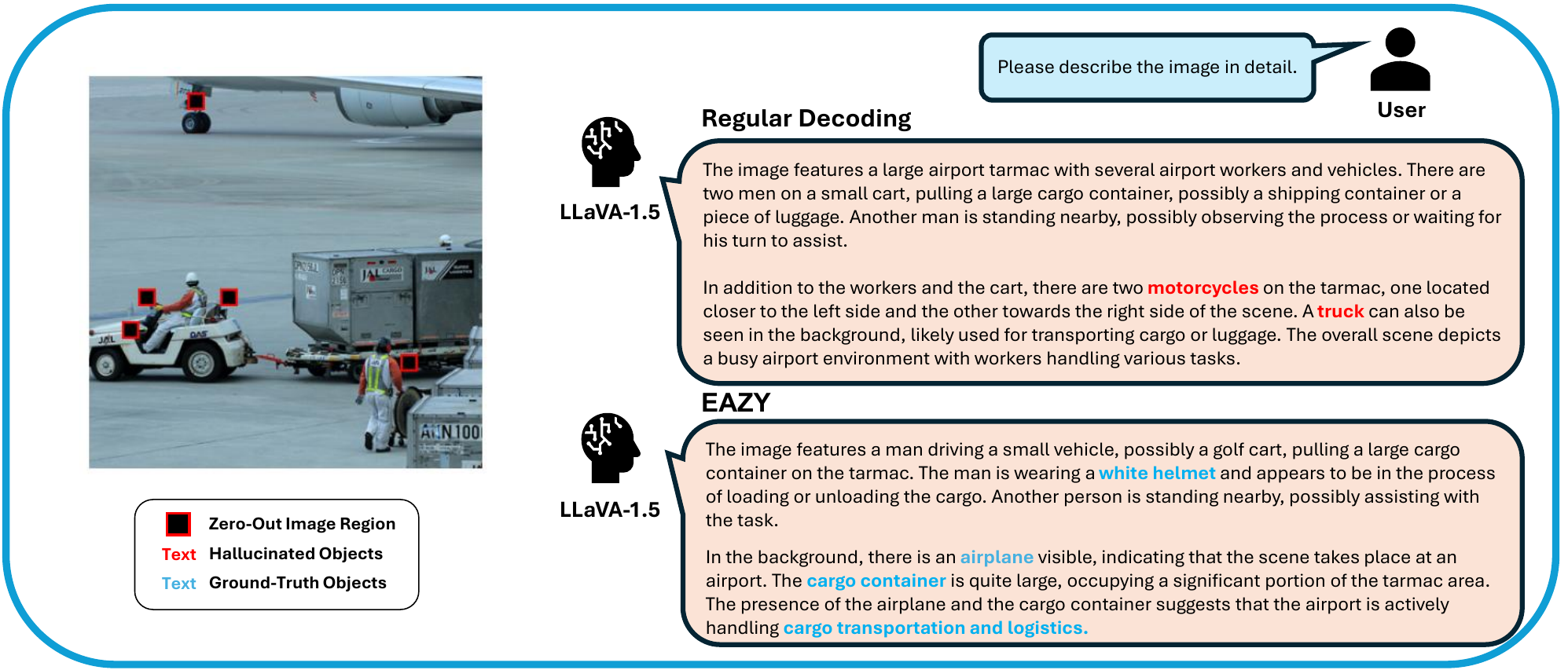}
    \caption{\textbf{Case 1:} When using original greedy decoding, the response generated by LLaVA-1.5 model contains "motorcycles" and "truck", two HOs. After using \eazy to remove five image tokens, it can be observed the HOs disappeared in the new response with more details and better description.}
    \label{fig:airport_case}
\end{figure*}

\begin{figure*}[!h]
    \centering
    \includegraphics[width=1\linewidth]{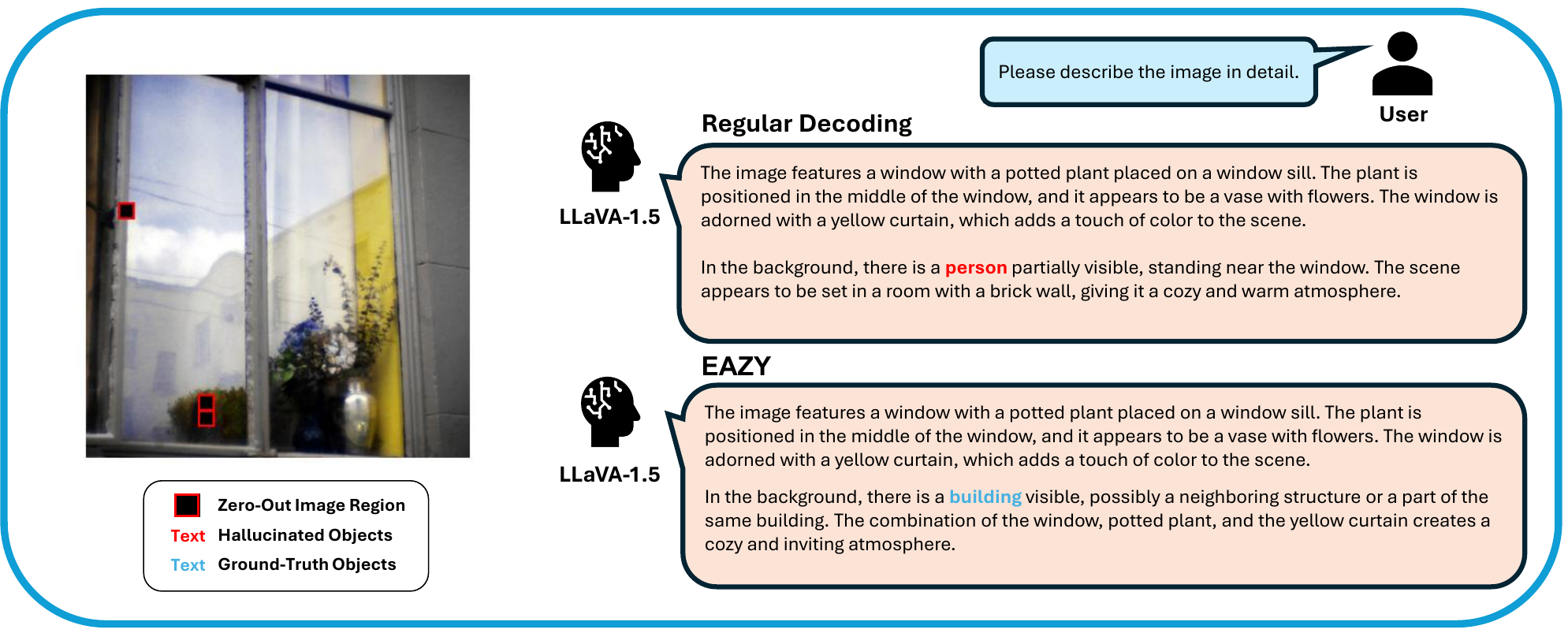}
    \caption{\textbf{Case 2:} The LLaVA-1.5 falsely recognized a nonexistent person in the image. \eazy identified the building reflected in the window.}
    \label{fig:vase}
\end{figure*}


\begin{figure*}[!h]
    \centering
    \includegraphics[width=1\linewidth]{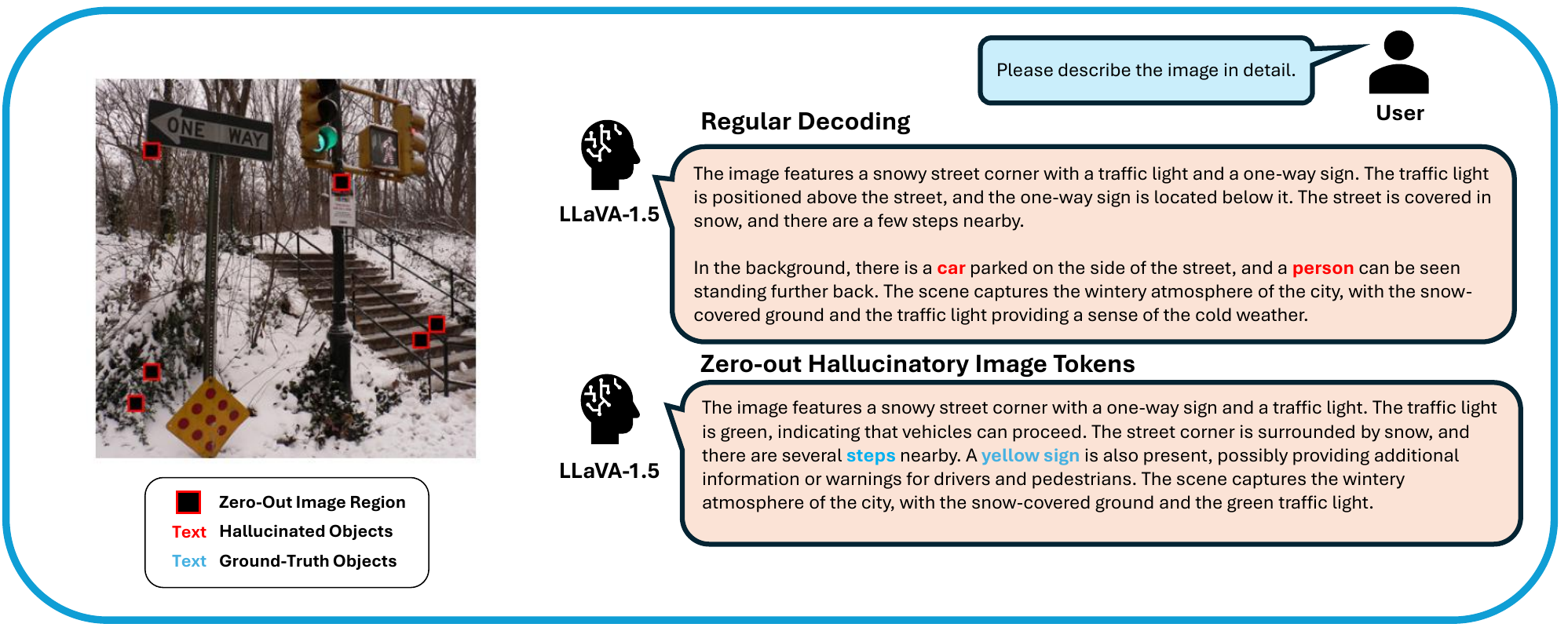}
    \caption{\textbf{Case 3:} In the original response, the model identified a car and a person in the image. \eazy successfully helped the model recognize the steps and the yellow markings while eliminating the previous hallucinations.}
    \label{fig:street_case}
\end{figure*}

\begin{figure*}[!h]
    \centering
    \includegraphics[width=1\linewidth]{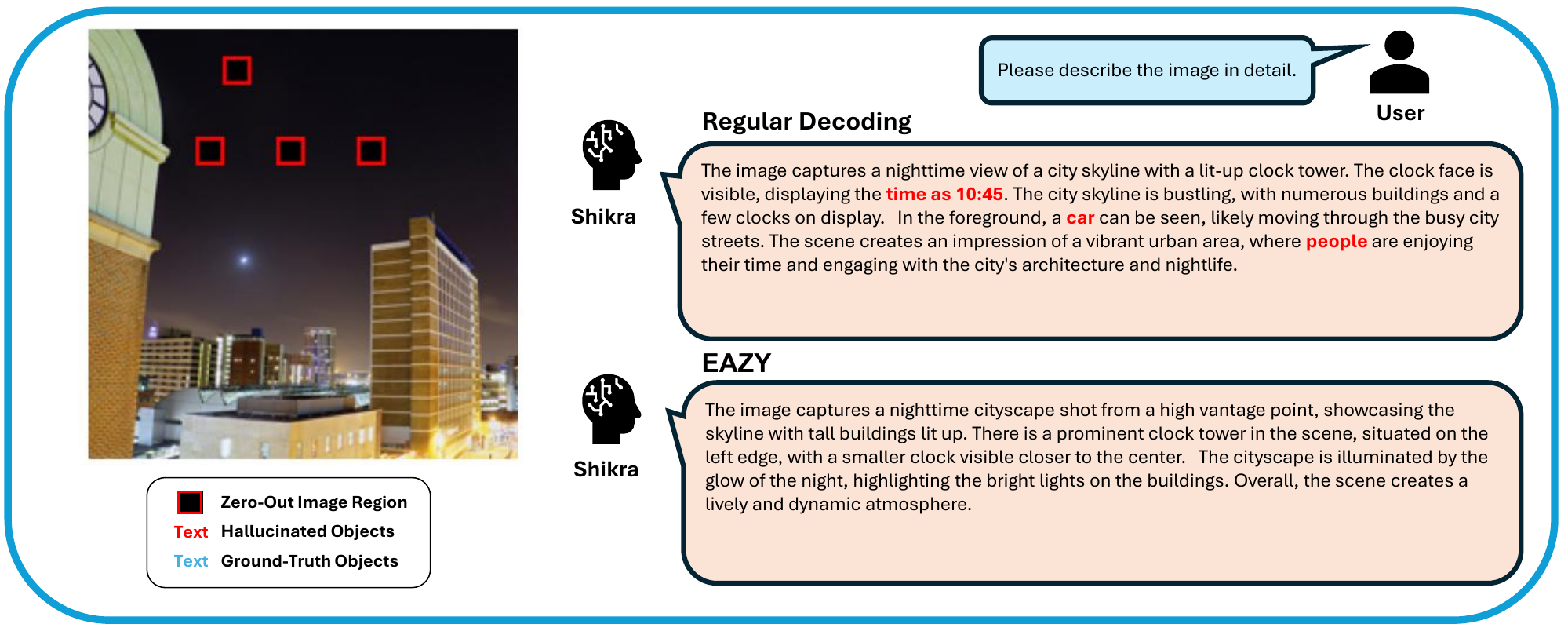}
    \caption{\textbf{Case 4:} \eazy not only removed the hallucinated objects \textit{(car} and \textit{people}) but also corrected the inaccurate clock time.}
    \label{fig:night_case}
\end{figure*}

\begin{figure*}[!h]
    \centering
    \includegraphics[width=1\linewidth]{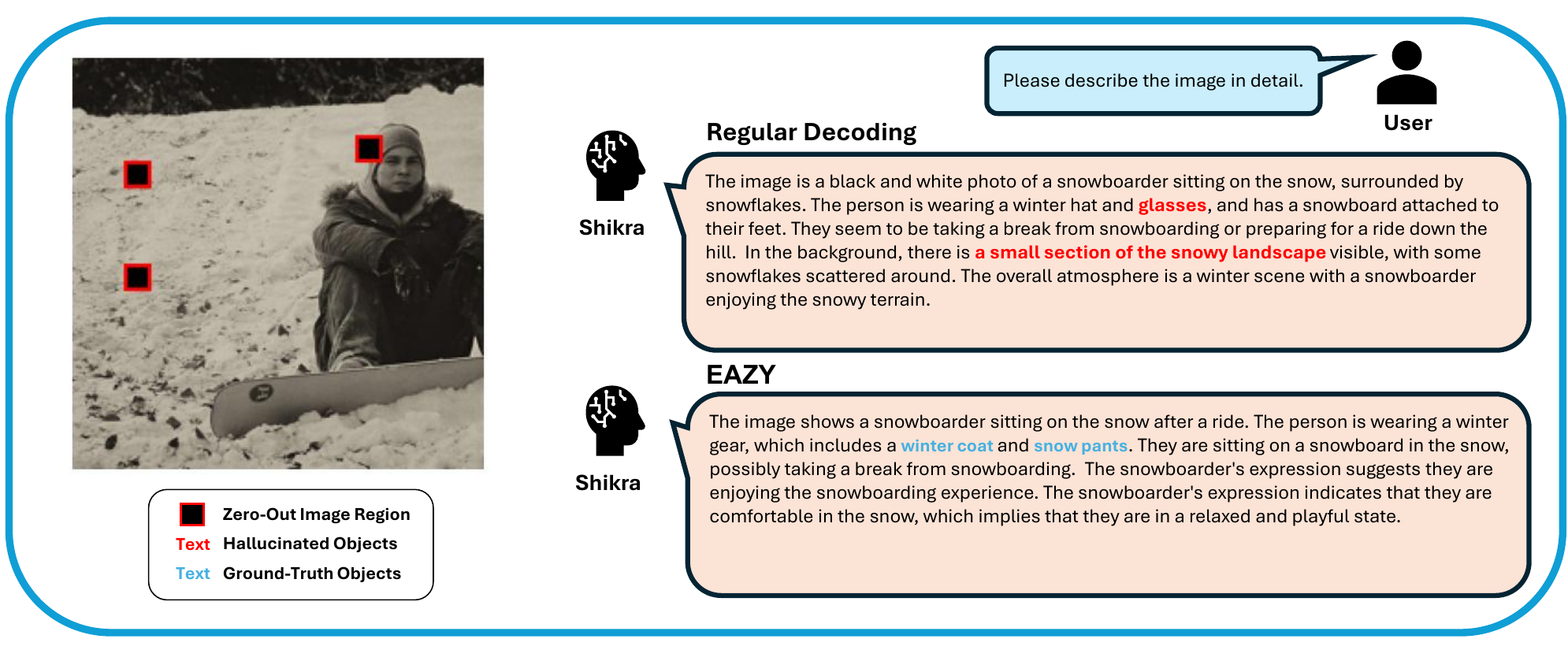}
    \caption{\textbf{Case 5:} \eazy corrected the hallucinated object (\textit{glasses}) and adjusted the inaccurate description of the background environment (\textit{a small section of the snowy landscape}).}
    \label{fig:snow_case}
\end{figure*}

\begin{figure*}[!h]
    \centering
    \includegraphics[width=1\linewidth]{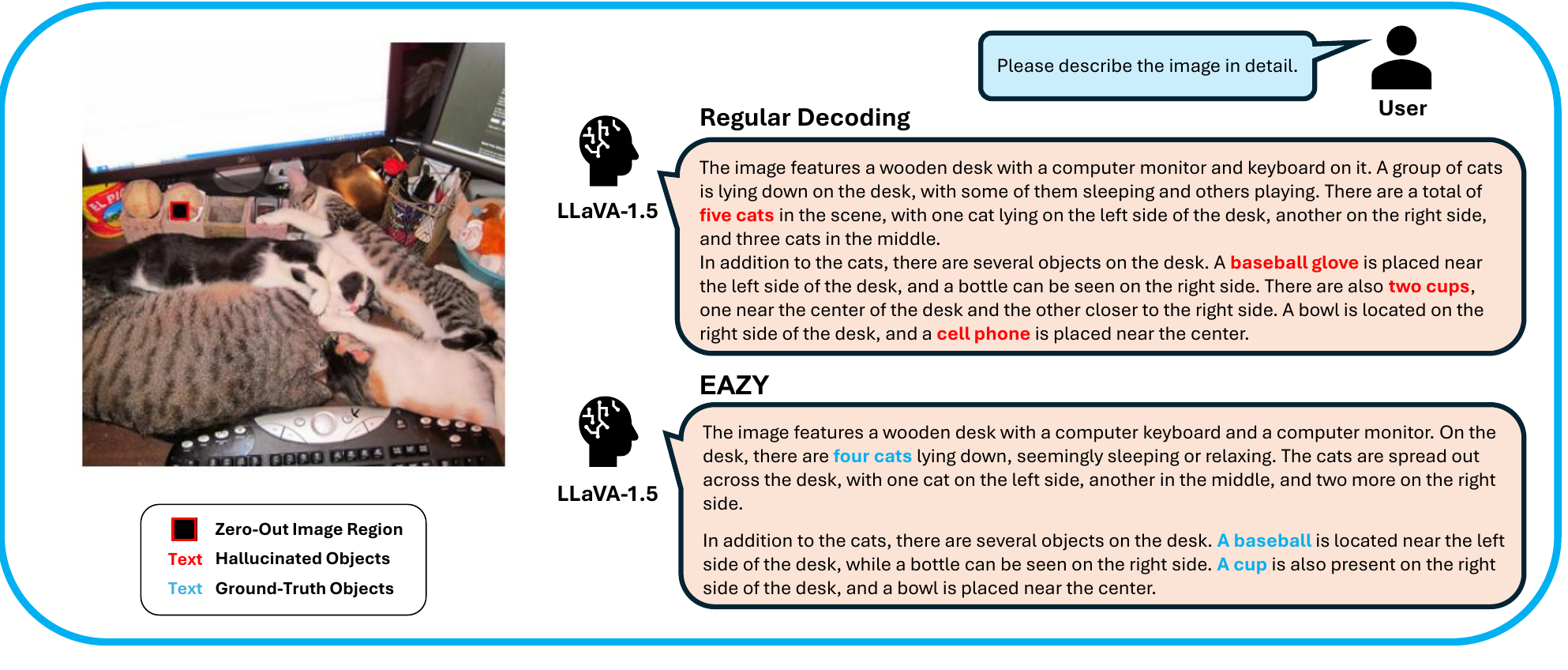}
    \caption{\textbf{Case 6:} By zeroing out just a single image token, the model successfully identified the correct number of cats and cups in the image while removing the hallucinated objects \textit{baseball glove} and \textit{cell phone}.}
    \label{fig:cat_case}
\end{figure*}

\begin{figure*}[!h]
    \centering
    \includegraphics[width=1\linewidth]{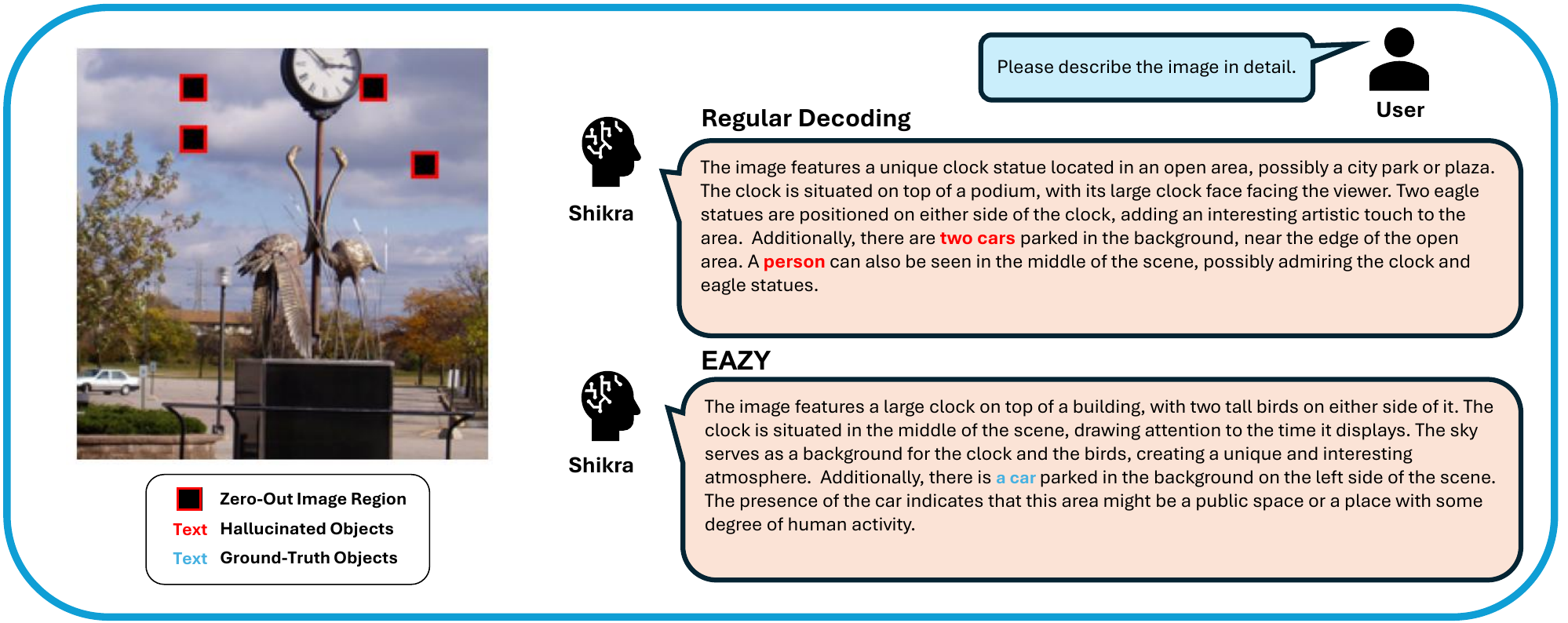}
    \caption{\textbf{Case 7:} With \eazy, the number of cars in the new response was correctly identified, and the hallucinated object \textit{person} was successfully removed.}
    \label{fig:square_case}
\end{figure*}

\end{document}